\documentclass{article}

\usepackage[nonatbib,final]{neurips_2022}

\usepackage[utf8]{inputenc} 
\usepackage[T1]{fontenc}    
\usepackage{url}            
\usepackage{booktabs}       
\usepackage{amsfonts}       
\usepackage{nicefrac}       
\usepackage{microtype}      
\usepackage{xcolor}         

\emergencystretch=\maxdimen
\usepackage{amsmath}
\usepackage{subcaption}
\usepackage{graphicx}
\usepackage{algorithm}
\usepackage{algpseudocode}

\title{Actively Learning Costly Reward Functions for Reinforcement Learning}

\author{
 André Eberhard$^1$\\andre.eberhard@kit.edu
 \And
 Houssam Metni$^2$\\houssam.metni@etu.unistra.fr
 \AND
 Georg Fahland$^1$\\georg.fahland@kit.edu
 \And
 Alexander Stroh$^1$\\alexander.stroh@kit.edu
 \And
 Pascal Friederich$^1$\\pascal.friederich@kit.edu
 \AND \\$^1$Karlsruhe Institute of Technology, $^2$Université de Strasbourg
}

\begin{document}

 \maketitle

 \begin{abstract}
 Transfer of recent advances in deep reinforcement learning to real-world
 applications is hindered by high data demands and thus low efficiency and
 scalability.
 Through independent improvements of components such as replay buffers or
 more stable learning algorithms, and through massively distributed systems,
 training time could be reduced from several days to several hours for standard
 benchmark tasks.
 However, while rewards in simulated environments are well-defined and easy
 to compute, reward evaluation becomes the bottleneck in many real-world
 environments, e.g., in molecular optimization tasks,  where computationally
 demanding simulations or even experiments are required to evaluate
 states and to quantify rewards.
 Therefore, training might become prohibitively expensive without an
 extensive amount of computational resources and time.
 We propose to alleviate this problem by replacing costly ground-truth
 rewards with rewards modeled by neural networks, counteracting
 non-stationarity of state and reward distributions during training with an
 active learning component.
 We demonstrate that using our proposed ACRL method (\textbf{a}ctively
 learning \textbf{c}ostly rewards for \textbf{r}einforcement
 \textbf{l}earning), it is possible to train agents in complex real-world
 environments orders of magnitudes faster.
 By enabling the application of reinforcement learning methods to new
 domains, we show that we can find interesting and non-trivial solutions to
 real-world optimization problems in chemistry,  materials science and
 engineering.
\end{abstract}

 \section{Introduction}\label{sec:introduction}
Reinforcement Learning (RL) techniques have achieved impressive results in a
wide range of applications such as
robotics~\cite{Kober13}, games~\cite{Mnih15,Silver16,Vinyals19} or natural
sciences~\cite{Mahmud18,Zhou19}.
This success is the result of improvements along multiple independent
branches of RL research such as an improved understanding of rewards in
difficult environments~\cite{Schaal97, Abbeel04, Christiano17, Wirth17}, more
sample-efficient training via experience
replay~\cite{lin-experience-replay, prioritized-experience-replay,
 hindsight-experience-replay, focused-experience-replay}
or more effective sampling via active
learning~\cite{daniel2015active, cui2018active, preference-based-gpr},
more powerful algorithms~\cite{Mnih15, ddqn} and more efficient and scalable
implementations~\cite{dist-prioritized-experience-replay,
 nvidia-gpu-simulators, rainbow} of established techniques.
These extensions were primarily developed and benchmarked in simulated
environments such as OpenAI Gym~\cite{openai-gym} or
MuJoCo~\cite{mujoco}, where rewards are well-defined and computationally
cheap to obtain.
However, in real-world tasks rewards may be either difficult to formulate or
to collect.
There has been extensive work on how to formulate and quantify rewards in
scenarios where agents have to learn from demonstrations~\cite{Schaal97,
 Abbeel04} or from ranked alternatives~\cite{Christiano17, Wirth17}.
These methods are mainly developed within the field of robotics, where
feedback frequently is provided by a human supervisor.
Thus, since human feedback is relatively expensive, it would be desirable to
reduce the number of expert evaluations.
Active reward learning techniques aim to reduce the number of expert queries
by selecting only the most informative ones, usually employing uncertainty
measures within the Bayesian
framework~\cite{daniel2015active, cui2018active, preference-based-gpr}
as a selection criterion.
More recently, \textit{Information Directed Reward Learning (IDRL)
}~\cite{inf-dire-reward-learning} has been proposed to learn an unknown
reward function with as few expert queries as possible.

Existing literature focuses on developments in simulated environments and
real-world tasks in fields such as robotics.
While in the former case rewards are clearly formulated and cheap to obtain,
in the latter case rewards are typically difficult to formulate and/or
quantify, e.g., in object manipulation tasks~\cite{cloth-manip}.
However, in many other fields rewards appear to have different properties
than in these scenarios, for which most of existing work has been done.
In a natural sciences and engineering context, for example, rewards are
frequently the result of computationally demanding optimization procedures
or algorithms.
Thus, it may not be possible to leverage recent advances in RL to make
training more efficient in these scenarios, since frequent reward evaluation
may become a major bottleneck during training without an extensive amount of
computational resources and horizontal scaling.

In this paper, we present a framework to make training of RL agents feasible
in environments where it is prohibitively expensive to evaluate ground-truth
rewards at every step.
We propose to use neural networks as reward function approximators with
active learning to address non-stationarity of state and reward
distributions during training.
We show that within our framework, which we term \textbf{ACRL}
\footnote{Our code is available at:
\url{https://github.com/32af3611/ai4mat-neurips-workshop-2022}},
neural networks, pre-trained on a relatively small initial dataset and
regularly updated during training via an \textbf{a}ctive learning approach,
can be used as reward proxies and that agents trained within this framework
achieve competitive results across different real-world tasks with varying
computational \textbf{c}ost, thus extending the applicability of \textbf{RL}
algorithms to a wide range of new applications.
This way, our agents are able to generalize in environments with varying
constraints which avoids re-optimizations for new instances of a task.
 \section{Related work}\label{sec:related-work}

\subsection{Learning reward functions}\label{subsec:reward-type}
In theory, every agent accumulates rewards under a unified mathematical
framework.
In practice, though, the exact properties of a reward function depend on the
task.
For example, rewards can be immediate or delayed and the reward signal can
be binary, discrete or real.
In simulated environments like OpenAI Gym and MuJoCo rewards are well-defined
and exposed to the agent via a simulator interface.
In fields like robotics, rewards can become complex, high-level signals of
desired behavior, e.g., to manipulate an object in a particular
manner~\cite{cloth-manip}.
Since the formulation of a reward function is often difficult in the latter
case, early work~\cite{Schaal97,Abbeel04} aimed to infer an unknown reward
function solely from demonstration.
While alleviating the issue of reward formulation, demonstrations by a human
supervisor are costly to obtain.
As an alternative, preference-based learning~\cite{Christiano17,Wirth17}
allows feedback to be a relative preference over a set of trajectories
rather than a quantitative measure of goodness.

In our work, we focus on problems where the reward function is known, e.g.,
as the solution of an optimization problem, but difficult to evaluate.
This is frequently the case in real-world applications, e.g., in natural
sciences and engineering, where the solution to a problem can be formulated
as a goal-directed search implemented as an agent's policy.
Thus, states of high rewards correspond to more optimal solution spaces of
the underlying problem.
We therefore propose to replace the known reward function with an
approximate model and to jointly train it with the agent to account for
non-stationarity of state and reward distributions during exploration.
Due to their ability to generalize, our agents are able to solve
optimization tasks with varying constraints, which is, in general, not
trivially doable using conventional optimization and search methods.

\subsection{Active reward learning}\label{subsec:active-reward-learning}
Active reward learning techniques~\cite{daniel2015active, cui2018active,
 preference-based-gpr} build upon the insight that not all training samples
are equally important for learning and aim to select only those samples
which are most beneficial for learning.
The selection is usually done by some form of uncertainty estimation, often
within the Bayesian framework.
Reducing the number of state queries is vital in cases where reward
evaluation is expensive.
While existing work employs active learning to reduce the number of queries
for the agent to accelerate convergence of the RL training, we employ active
learning for the reward model such that predictions become more accurate on
states the agent visits during exploration.
Closest to our method is \textit{Information Directed Reward Learning (IDRL)
}~\cite{inf-dire-reward-learning}, which also uses a reward model and
active reward learning to accelerate training.
Yet, there are important conceptual differences to our method:
\begin{itemize}
 \item \textit{IDRL} assumes the absence of a reward function, while in our context the reward function is known.
 \item \textit{IDRL} makes learning more efficient w.r.t.\ the number of training steps,
 our method makes
 training more efficient w.r.t.\ wall-clock time.
 \item \textit{IDRL} uses active learning for faster convergence, we use active learning
 for the exploration
 of relevant solution spaces.
 \item \textit{IDRL} relies on the Bayesian framework for uncertainty estimation, while
 our method requires standard deep learning components only.
 \item \textit{IDRL} trains multiple policies for query selection which does not scale to
 tasks where agents
 are expensive to train.
 In contrast, our method trains only one agent.
\end{itemize}

\subsection{Sample-efficiency}\label{subsec:sample-efficiency}
Active learning techniques improve sample-efficiency in terms of sample
collection.
In vanilla RL, every observation is used only once to update the agent's
policy, making learning slow and sample-inefficient.
A popular technique to overcome this is to use \textit{experience
replay}~\cite{lin-experience-replay}, which improves sample-efficiency in
terms of sample usage by storing experience in a \textit{replay buffer} and
performing parameter updates on batches uniformly sampled from it.
Improvements of experience replay use different forms of non-uniform
sampling~\cite{prioritized-experience-replay, focused-experience-replay},
handle sparse and binary reward signals and multi-goal
environments~\cite{hindsight-experience-replay},
and are also extended to a distributed
context~\cite{dist-prioritized-experience-replay}.

In our work, we do not aim to increase sample-efficiency of the RL training
process.
Rather, we avoid expensive ground-truth evaluations for known regions of the
state space by using a reward model.
We increase the size of this region over the course of training by providing
ground-truth labels for a small fraction of states selected by some sampling
method.

\subsection{Efficient implementations}\label{subsec:efficient-implementations}
The effects of other extensions within the RL framework have been studied
in~\cite{rainbow}, showing recent advances can be integrated to improve
their standalone-performance.
From a practical point of view, the authors of~\cite{acc-methods-drl}
provide a unified implementation view of RL algorithms to leverage modern,
parallel hardware architectures to further reduce training time.

In our work, we do not aim to improve RL from a technical perspective.
Rather, we propose an extension to restore the effectiveness of these
methods in scenarios where their efficiency would be threatened by the
reward evaluation bottleneck.
We note that our method is scalable and naturally can be integrated into
distributed architectures such as~\cite{dist-prioritized-experience-replay}.
 \section{Our method: ACRL}\label{sec:our-method}
Existing literature covers how to learn a reward function in cases of
unclear tasks or how to make efficient use of it in cases where it exists
and can be evaluated frequently.
In contrast to that, in many other tasks the reward function is clearly
defined but costly to evaluate.
Providing these kinds of rewards to an agent during training thus can become
prohibitively expensive even with off-policy learning with experience replay
as one may fail to gather enough examples to learn from.
In the following we describe our proposed ACRL framework to alleviate this
issue.
We use a standard MDP formulation as found in~\cite{sutton2018reinforcement}.

Let $f(s)$ be a quantity or metric associated with state $s$, $f$ being a
known but expensive to evaluate function of $s$.
Without loss of generality, we aim to find a (local) minimum of $f$, or
equivalently, a (locally) optimal
state $s^{*} = \underset{s}{\arg\min}\;f(s)$.
Due to high computational cost as well as non-convexity of $f$ in real-world
tasks, we neither can directly solve for $s^{*}$ nor is it likely that we
can find $s^{*}$ with heuristic search in general.
We therefore propose a more principled search of $s^{*}$ by framing it as a
sequential decision-making problem within the RL framework.
A natural definition of reward in such environments is $r_{t} = f(s_{t-1}) -
f(s_{t})$, i.e., the agent aims to accumulate reward by sequentially
visiting states $s$ with decreasing $f(s)$.
Let $s_{0}$ be a possibly random initial state, the agent then aims to
maximize the total cumulative reward
$R_{T} = \sum_{t=1}^{T} f(s_{t-1}) - f(s_{t}) = f(s_{0}) - f(s_{T})$.
Due to the computational complexity of $f$, training an agent for a large
number of steps may become infeasible or at least very time-consuming.
To reduce the computational burden of state evaluations during training, our
framework requires only a few modifications of the standard training loop,
namely the introduction of a reward model and its improvement via active
learning.
The two steps are then as follows.

The first step is to pre-train an approximate reward model 
$\hat{f}$, e.g., a neural network, on a small, initial dataset $D$
in a supervised manner.
$\hat{f}$ is then used as a drop-in replacement for the true
evaluation function $f$.
Doing so is theoretically sound as the reward distribution does not depend on the agent's policy.
This allows using our framework with both value-based and policy gradient methods without the necessity to change the underlying theory.
At this point, we make several mild assumptions about $f$.
In contrast to the general RL setting, we assume that we can
evaluate $f$ in any state, thus providing dense and instantaneous
rewards on state transitions.
Hence, our method is not well-suited for sparse or delayed rewards.

The second step is then to actively improve the reward model 
during agent training.
Since the initial state distribution in $D$ likely differs from states
visited by an exploring agent, $\hat{f}$ may have poor extrapolation
capabilities which will cause agent training to diverge as estimated
state quantities may not have their true value predicted accurately.
This particularly applies in scenarios where it is difficult to define
\textit{good} initial states, for example in the case of optimization
problems where the optimal solution is to be found rather than given.
To overcome this issue, we propose to sample a small number of states
encountered during agent training and to provide the expensive 
ground-truth labels for them.
In the most general form, we define an acquisition function $h(s)$ which 
hypothesizes about how beneficial adding the true label $f(s)$ to $D$ is for
training the reward model.
We then periodically evaluate $h$ for a small fraction of the agent's 
experience $\mathcal{E}$, e.g., the last $N$ steps, where $N$ is an application-dependent hyperparameter.
We set $s' = \underset{s \in \mathcal{E}}{\arg\max}\;h(s)$, $D = D \cup \{s'\}$ and subsequently update $\hat{f}$ on the new $D$, either by training from scratch or fine-tuning.
At this point, we assume that reward model can be trained reasonably fast such that the training time can be amortized given enough reward evaluations.
For example, $h(s)$ may be chosen to be $\hat{f}(s)$, $||\nabla\hat{f}(s)||$ or other sampling techniques like uniform
or uncertainty sampling.
We hypothesize that this active learning component allows to explore
relevant regions of the state space effectively and efficiently.
An important implication of using active learning is that, depending on the
task at hand, the initial reward model must not be perfectly accurate.
For example, when using our method for optimization tasks where it is
unlikely that the optimal solution space is included in the initial dataset
$D$, perfect accuracy in this space is not necessary since the agent moves
away from the initially covered space towards a more optimal region.
It is thus more important for the reward model to be accurate on the 
on-policy distribution of states rather than on randomly selected initial
data points.
The reward model is only required to improve as the agent's policy improves
and stabilizes.
We found this active learning component to be crucial in our tasks.

A summary of the overall procedure can be found in
Algorithm~\ref{alg:approx-rewards}.
We note that even though we use variations of Double DQN~\cite{ddqn} agents
in all experiments, our method does not assume any particular type of RL
implementation and can be integrated into existing implementations with
minimal changes, even in asynchronous and distributed settings.

\begin{algorithm}
 \caption{Double Deep-Q-Learning within ACRL}
 \label{alg:cap}
 \begin{algorithmic}[1]
  \State agent $A$, replay buffer $B$, initial dataset $D$, environment $E$, reward network $\hat{f}$ trained on $D$

  \State $\hat{f}$ $\leftarrow$ train($\hat{f}, D$) \Comment train reward
  network
  \State E.reward $\leftarrow \hat{f}$ \Comment E.step() uses $\hat{f}$
  instead of $f$

  \For {episode = 1 to M}
   \Comment training loop
   \State $s_{t} \leftarrow$ initial state
   \For{step = 1 to T}
    \Comment episode loop
    \State $a_{t} \leftarrow$ A.action($s_{t}$) \Comment{$\epsilon$-greedy}
    \State $s_{t+1}, \hat{r}_{t+1} \leftarrow$ E.step($a_{t}$)
    \Comment{$\hat{r}_{t+1}=\hat{f}(s_{t}, a_{t})$}
    \State $obs \leftarrow$ ($s_t, a_t, \hat{r}_{t+1}, s_{t+1}$)
    \State B.add($obs$) \Comment{save observation}
    \State $obs \leftarrow$ B.sample() \Comment{sample experience from B}
    \State A.optimize($obs$) \Comment{update parameters}
   \EndFor

   \If {sample state}
    \Comment e.g., periodically
    \State $s' \leftarrow \underset{s \in \mathcal{E}}{\arg\max}\;h(s)$ \Comment any method
    \State $y' \leftarrow f(s')$ \Comment{calculate ground-truth label}

    \State $D \leftarrow D \cup \{(s',y')\}$
   \EndIf

   \If {update model}
    \Comment e.g., periodically

    \State $\hat{f}$ $\leftarrow$ train($\hat{f}, D$) \Comment retrain
    reward network
    \State E.reward $\leftarrow \hat{f}$ \Comment update reward network
   \EndIf
  \EndFor
 \end{algorithmic}
 \label{alg:approx-rewards}
\end{algorithm}

 \section{Applications}\label{sec:applications}

\subsection{Proof-of-principle: Molecular property
optimization}\label{subsec:proof-of-principle:-molecular-property-optimization}
The algorithm described above is first used in molecular property
optimization tasks as a proof-of-principle.
We use two fast-to-evaluate benchmarking properties to evaluate the
performance of the algorithm and to choose its optimal hyperparameters.
Both the Q-network and the reward network are trained on
Morgan fingerprint vectors as molecular
representations~\cite{Morgan65, Rogers10}.
States and actions are based on prior work~\cite{Zhou19}, where states
are discrete molecular graphs and actions are semantically allowed local
graph modifications.

The first benchmarking property is the penalized logP score, a widely used
metric in the literature for evaluating and benchmarking machine learning
models on regression and generative
tasks~\cite{geneticalgorithm, G_mez_Bombarelli_2018,gcpn}.
The logP score is the logarithm of the water-octanol partition coefficient,
quantifying the lipophilicity or hydrophobicity of a molecule.
Penalized logP additionally takes into account the synthetic accessibility
(SA) and the number of long cycles ($n_{cycles}$):
\begin{equation}
 pen.\,logP = logP - SA - n_{cycles}   
\end{equation}
The second benchmarking property used here is the QED score, which is a
quantitative estimate of druglikeness based on the concept of
desirability~\cite{Bickerton12}.
QED is an empirical score quantifying how "drug-like" a molecule is.
Both properties are computationally inexpensive and can be calculated using
RDKit~\cite{rdkit06}.
We use them as benchmarking properties to study the effect of replacing the
ground-truth reward with an approximation and to choose hyperparameters of
our algorithm.
In both applications, empty initial states are optimized for $T=40$ steps.
We then test our method on a real-life application in molecular improvement
with a more costly property value to calculate.

\subsection{Application I: Molecular
design}\label{subsec:application-i:-molecular-design}
In our first application, we evaluate ACRL on a molecular design task
involving more costly rewards.
We aim to optimize electronic properties of molecules such as energies of
the Highest Occupied Molecular Orbital~(HOMO) and the Lowest Unoccupied
Molecular Orbital~(LUMO) by performing sequential modifications.
These values can be calculated using semiempirical quantum mechanical
methods such as density functional tight binding methods as implemented in
\textit{xTB}~\cite{grimme2017,bannwarth2020}.
xTB-based reward evaluations on one Intel Xeon Gold 6248 CPU range from
seconds to minutes, depending on size and structure of the molecule.
Compared to other RL applications, this is comparably expensive, especially
considering the number of reward evaluations needed during agent training.
The algorithm described above is applied using the hyperparameters found
in the experiments of 
Section~\ref{subsec:proof-of-principle:-molecular-property-optimization}.
Here, the agent learns a more application-oriented optimization goal, i.e.,
how to decrease the LUMO energy of randomly sampled starting molecules with
only $T=5$ steps per episode, while keeping the HOMO-LUMO gap constant.
Therefore, the goal of the agent is to find optimal local improvements of
given molecules with a limited number of actions, i.e., changes of the
chemical structure.
Let $s_{0}$ be a randomly sampled molecule at the beginning of an episode,
then the improvement of the molecule $s_{t}$ at timestep $t$ over $s_{0}$ is
defined as:
\begin{equation}
 R(s_{t}) = -|\text{gap}(s_{t})-\text{gap}(s_{0})| - (\text{LUMO}(s_{t})
 -\text{LUMO}(s_{0}))\label{eq:equation}
\end{equation}
with $\text{gap}(s)=\text{LUMO}(s)-\text{HOMO}(s)$ being the HOMO-LUMO
energy difference of molecule $s$.

\subsection{Application II: Optimization of airflow drag around an
airfoil}\label{subsec:application-ii:-optimization-of-airflow-drag-around-an-airfoil}
The control technique of wall-normal blowing or/and suction constitutes a
promising approach for the reduction of drag in turbulent boundary
layers~\cite{Kinney.1967}.
This technique has been successfully utilized not only in flat-plate
boundary layers~\cite{Kametani.2011} but also on more complex curved
geometries like airfoils~\cite{Atzori.2020}.
The majority of studies on the aforementioned control technique, however,
considers uniform distribution of the introduced blowing or suction profiles.
In our second application, we use ACRL to minimize aerodynamic drag around
an airfoil by sequential adjustment of a set of blowing and suction
coefficients represented as vectors in $\mathbb{R}^d$ (see
figure~\ref{fig:profile}), which form the state space in $\mathbb{R}^{2d}$.
As higher coefficients trivially reduce drag, we seek to optimize profiles
with a constrained mean value for each side.
By choosing a different constraint at the start of each episode, we aim to
generalize across multiple instances of optimization.
We use a Double DQN~\cite{ddqn} agent with discrete actions corresponding
to exactly one (or no) modification of an entry of $s$ per step to keep the
action space as small as possible.
Thus, we seek to find a (near-)optimal state $s^{*} \in \mathbb{R}^{2d}$
under given constraints.
In our experiments, we use an episode length of $T=30$ steps.
While policy methods would be a more appropriate for this task, we use
Double DQN for the sake of consistency.

Let $d_{0} = f(s_{0})$ be the drag coefficient of starting state $s_{0}$
corresponding to a uniform profile on each side.
Our agent then seeks to find a sequence of modifications such that $R_{T} =
\sum_{t=1}^{T} d_{t-1} - d_{t} = d_{0}-d_{T}$ becomes as large as possible.
We note that while the agent seeks to maximize $R_{T}$, we are primarily
interested in the shape of states $s_{T}$ close to the (globally) optimal
state $s^{*}$ rather than the exact value of $f(s_{T})$.

The incompressible flow around airfoils is analysed using Reynolds-averaged
Navier–Stokes equation based simulations in order to assess the effect of
localized blowing and suction on the global aerodynamic performance of the
airfoil.
The simulations are carried out with the open-source CFD-toolbox
OpenFOAM~\cite{Weller.2011} using a steady state, incompressible solver.
For the current study we consider a flow around the NACA4412 airfoil at the
Reynolds number $Re=U_\infty c / \nu=4 \cdot 10^5$ and the angle of attack
$\alpha=5^\circ$.
For a more detailed description of the setup the reader is referred
to~\cite{Fahland.2021}.

One particular difficulty in training an RL agent in this scenario is the
fact that the true state evaluation function $f$ is a Computational Fluid
Dynamics (CFD) simulation.
On one core of an Intel~Xeon~Platinum~8368 CPU, the simulation runs for
approximately 10 minutes.
Due to a fixed mesh size, we found that parallelization beyond 4 cores did
not result in a significant speed-up, hence one reward evaluation takes
approximately 2 to 3 minutes and cannot be reduced significantly, which
severely limits the applicability of conventional RL algorithms with
thousands of sequential reward evaluations.

 \section{Results and discussion}\label{sec:results}

\subsection{Molecular property
optimization}\label{subsec:molecular-property-optimization}

\begin{figure*}[!htb]
 \centering
 \begin{subfigure}[b]{0.32\textwidth}
  \includegraphics[width=\textwidth]{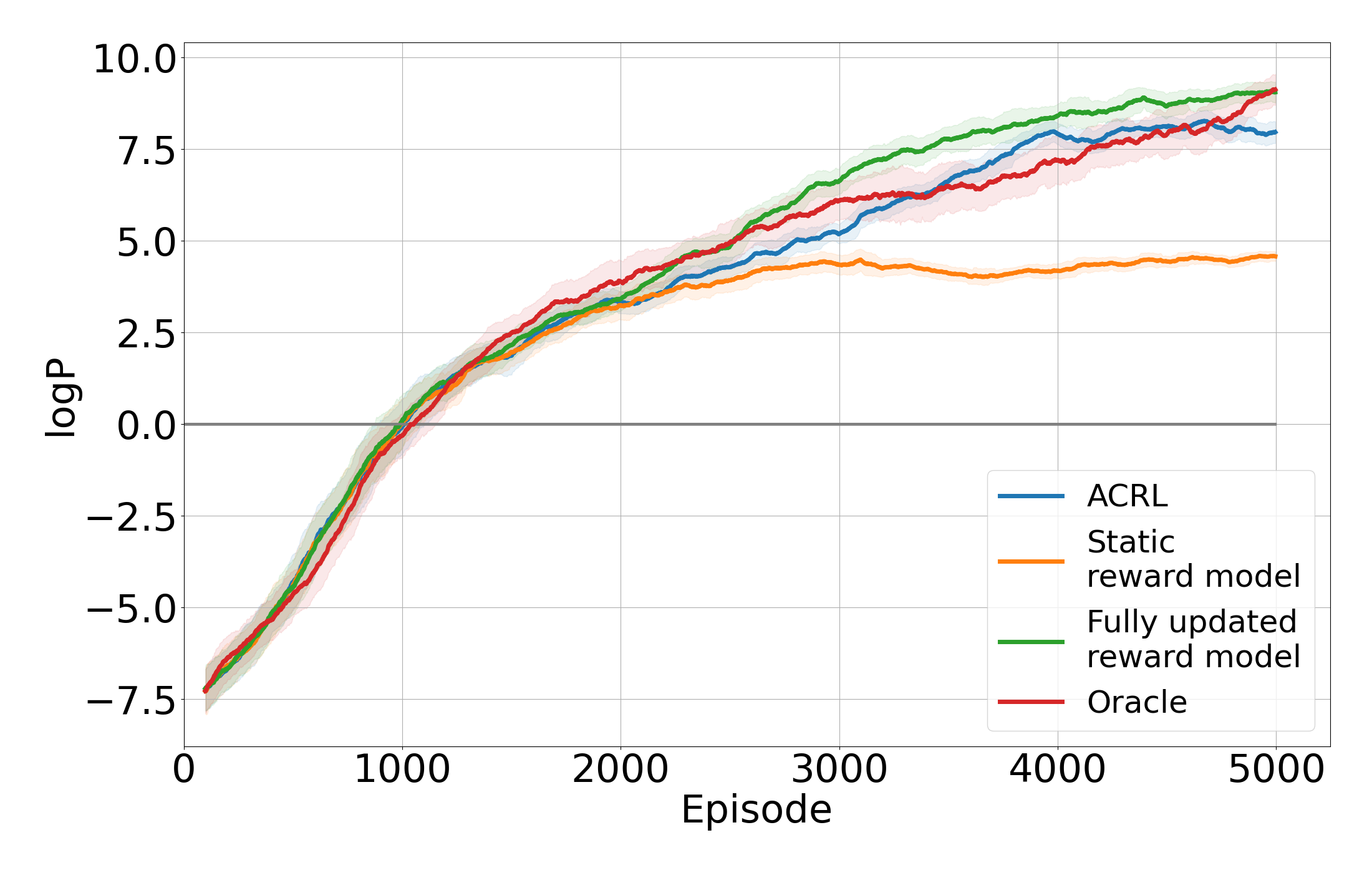}
  \caption{}
  \label{fig:logp}
 \end{subfigure}
 \hfill
 \begin{subfigure}[b]{0.32\textwidth}
  \includegraphics[width=\textwidth]{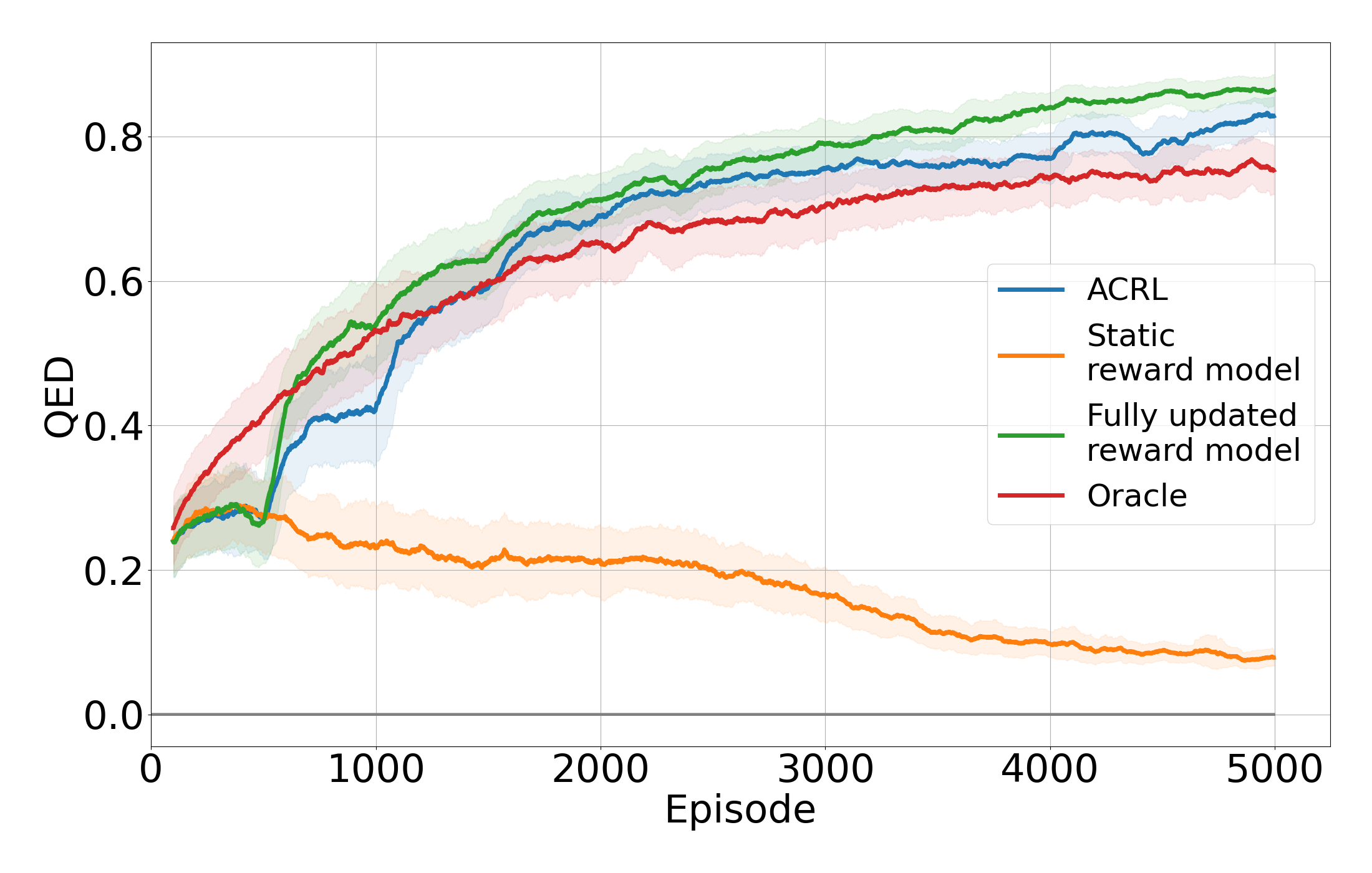}
  \caption{}
  \label{fig:qed}
 \end{subfigure}
 \hfill
 \begin{subfigure}[b]{0.32\textwidth}
  \includegraphics[width=\textwidth]{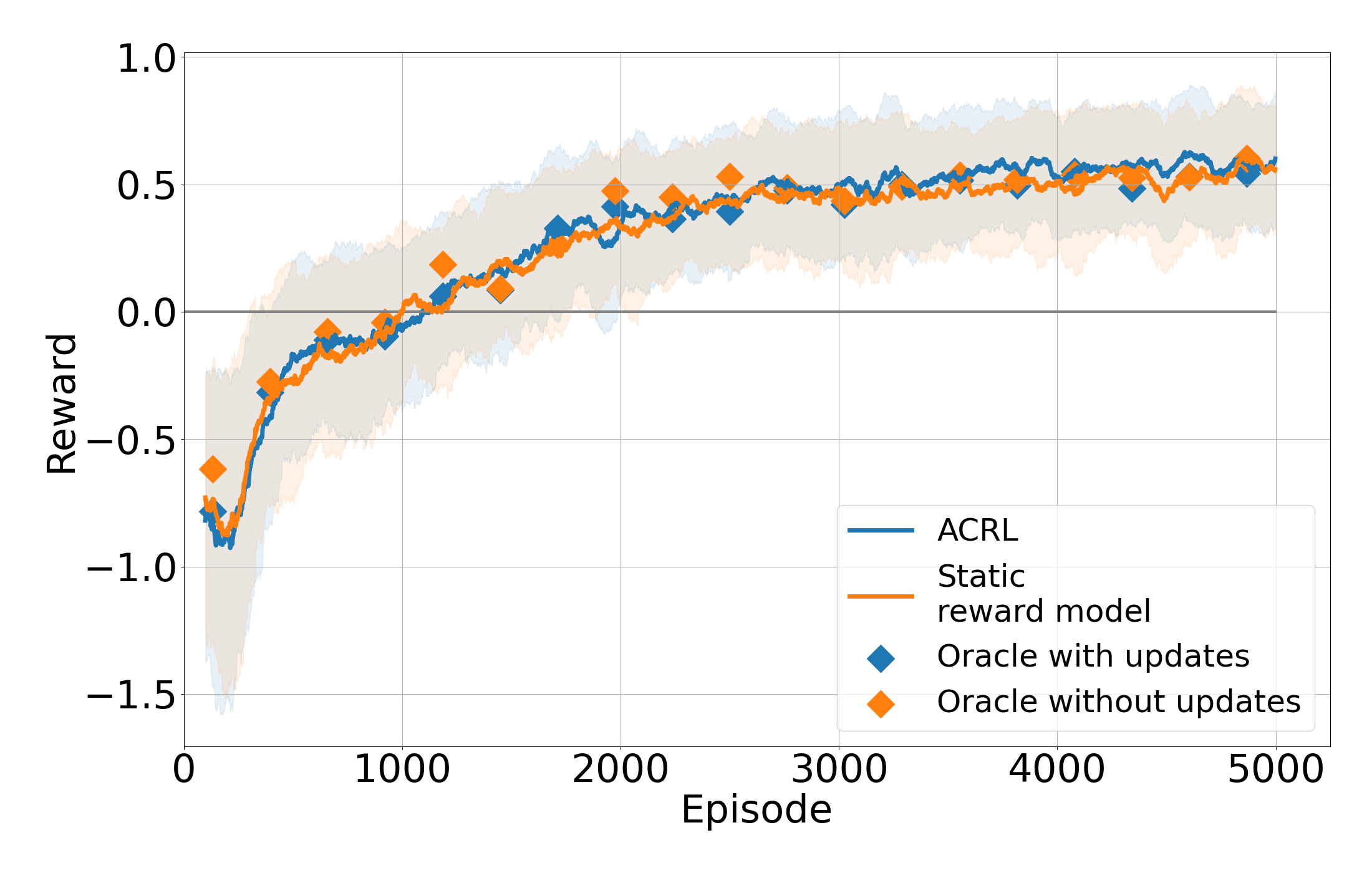}
  \caption{}
  \label{fig:mol_opt}
 \end{subfigure}
 \caption{
  Evolution of the reward reached by the agent during the optimization of
  logP and QED: The red curve was obtained by training the agent on real
  (oracle-based) rewards, while the blue, orange and green curves are the
  ACRL model,   the static reward model and a fully updated reward model,
  respectively.
  Due to high computational costs, only ACRL and static models can be tested
  in (c).
 }
\end{figure*}

Based on prior work by Zhou {\textit{et al.} }\cite{Zhou19}, we used cheap
chemistry benchmarking properties logP and QED as a proof of concept to
evaluate how the use of actively learned rewards performs in comparison to
the real reward.
Figure~\ref{fig:logp} and~\ref{fig:qed} show the performance of three
different agents with NN-approximated rewards compared to a reference agent
("oracle-based reward") trained on the real reward.
One of the reward approximation agents is only trained once in the beginning
("static").
One of the agents ("ACLR") uses a reward model which is updated at regular
intervals using additional oracle queries selected based on uncertainty
sampling.
The last agent ("full update") is updated after every episode using oracle
queries of all states encountered in that episode (i.e., closest to the
reference agent which directly uses oracle queries for training).
After approximately 2000 episodes in case of logP optimization and already at
the beginning of QED optimization, the performances of the agents start to
differ.
While the performance of the static agent stagnates, all three other agents
show similar performance.

The failure of the \textbf{static agent} to learn is due to the low
generalization ability of the initial reward model itself, which is trained
on the QM9 dataset~\cite{Ruddigkeit12,Ramakrishnan14} containing
approximately 134.000 molecules with up to 9 non-hydrogen atoms.
To some extent, the weak generalization can be attributed to not using
state-of-the-art graph neural networks.
However, we decided to use the same molecular representation and model as in
the original Q-networks in Zhou {\textit{et al.}}\cite{Zhou19}, i.e.,
fingerprint representations and MLPs.
Furthermore, during the learning process, the molecules generated
(especially after a high number of episodes) contain many more atoms, which
explains why the static reward model fails to correctly estimate the real
property values.
The strength of this effect depends on the property studied.
In the logP optimization task, the property values reached with a static
reward model follow the general trend of learning with real reward, even
though the final performance after 5000 episodes is lower.
In case of QED optimization, the static reward model fails to predict
QED-values for molecules outside the training distribution.
As a consequence, the RL agent learns to exploit errors of the static reward
model and finds adversarial examples, rather than samples with desirable
properties.

The active learning component within \textbf{ACRL agent} allows the reward
model to learn from molecules outside its initial training distribution,
thus improving reward evaluations during agent training.
By only selecting a small subset of labels obtained using oracle queries to
be added to the training set, the objective of the ACLR agent is to mimic
the reference agent's real behavior as closely as possible.
This includes finding (nearly) optimal points (see e.g.,
~\cite{inf-dire-reward-learning}) to be selected for retraining of the
reward model to minimize its errors while at the same time minimizing the
number of costly oracle queries.
We experimented with different sampling strategies (see SI), from which a
query-of-committee model~(see~\cite{10.1145/130385.130417}) performed best.
Therefore, in the ACLR model used in the molecular design and improvement
tasks, three reward models were trained independently to form a
query-of-committee model.
The three reward models are retrained after 500 episodes with the initial
training set along with all 400 new molecules generated during the agent
learning process and their computed real property values.
The selection of new oracle queries to extend the dataset is based on the
disagreement between the three reward models measured by the standard deviation of the predictions.
However, our work is independent of the particular sampling strategy (even
random sampling of visited states can work well in some applications), as
long as the reward model's training distribution follows the exploration of
the RL agent.
Overall, the speed-up achieved by the ACRL model in this experiment compared
to the fully updated and oracle-based model is $50$ (see
Table~\ref{tab:speedup}).
The relationship between speed-up and rewards reached is analyzed in the SI\@.

\textbf{Fully updating} the reward model on oracle queries of all samples
("full update") aids the learning process.
In case of logP optimization and even stronger in case of QED optimization,
learning by fully updating the reward model has even surpassed learning with
actual reward values at certain episodes.
One potential reason for that can be that the exploration of the fully
updated agent is stronger than that of the reference agent (see SI), which
needs to be confirmed in future work.
However, in practice, fully updating the reward model by adding every single
generated point (along with its real property value) to the initial dataset
and retraining the neural network is as expensive as training the reference
agent, so it cannot be applied to tasks with costly rewards.

In order to understand why the fully updated reward model in some cases (e.g.,
Figure~\ref{fig:qed}) outperforms the oracle-based training, we analyzed the
effect of additional noise and thus exploration which might be induced by
replacing oracle-based rewards with (noisy) approximated rewards.
We therefore varied the $\epsilon$-greedy strategy of the learning process.
In particular, we varied final $\epsilon$ values (i.e., probabilities of
random actions) and the form of the $\epsilon$-decay function used on the
learning process.
However, none of the changes in $\epsilon$-decay could improve the learning
behaviour, i.e., the $\epsilon$-decay rate and function used
by~\cite{Zhou19} was optimal.
Therefore, for the rest of the simulations we used a fully exponential decay
reaching approximately 1\% randomness in episode 5000.
The results of this study are available in the supplementary information
section.
Further study of the improvement effect due to a fully updated reward model
is part of ongoing work as it has the potential to improve the performance
of RL agents with little computational overhead.

\subsection{Molecular improvement}\label{subsec:molecular-improvement}

After evaluating the performance of our agent on easy-to-compute properties
such as penalized logP and QED, we test our ACRL approach on a molecular
improvement task with more costly rewards, where an oracle-based reference
study is unfeasible.
In particular, we study a RL agent with the goal of independently varying
two quantum mechanically calculated energy levels of molecules with only
very few, in our case five, modification steps (see
Section~\ref{sec:applications}).
Figure~\ref{fig:mol_opt} shows the evolution of the ACRL and the static
reward agents' rewards as a function of the training episode.
We observe that the reward becomes positive after approximately 1000
episodes and stagnates after approximately 2000 episodes.
Therefore, the agent has learned to improve given (arbitrary) molecules,
since the reward value of the starting reference molecule is zero, each
episode starts with a randomly sampled molecule, and any molecule with
negative reward would have less desirable properties than the initial one.
This suggests that even though the agent deals with different starting
reference molecules at each episode, it has managed to learn a strategy to
increase the reward in a limited number of steps.

In contrast to the property optimization task discussed before, the
performance of the ACRL and the static reward agents are equal within the
confidence intervals.
A likely explanation for this observation is that the number of steps per
epoch in this task is limited to five, whereas 40 steps were possible in the
prior task.
Therefore, the agent here cannot generate molecules that are far outside the
initial distribution of starting molecules, i.e., the QM9 dataset.
Furthermore, the ratio of reward model queries to oracle queries in this
experiment is comparably low (see Table~\ref{tab:speedup}), meaning that the
ACRL reward model is updated on a high fraction of actually encountered
molecules.
A reference calculation with oracle based rewards or a fully updated reward
model to check if the ACRL model found near-optimal results (within the DQN
framework) are computationally too costly here and thus unfeasible.
However, we compared the predictions of the reward models for randomly
selected molecules throughout the training process to oracle predictions
(see points in Figure~\ref{fig:mol_opt}).
We found excellent agreement, indicating that the ACRL as well as the static
reward models are reliable.
Thus, the solutions found are not exploiting weaknesses of the reward
models, nor is the training limited by wrong predictions of the reward models.
Therefore, it is likely that the found solutions are of comparable quality
as ones that a hypothetical oracle-based RL model would find.
The speed-up achieved in this experiment compared to a hypothetical oracle
based model is $6.25$, which still has room for improvement,
given the high reliability of the reward models.

\subsection{Optimization of airflow drag around an
airfoil}\label{subsec:optimization-of-airflow-drag-around-an-airfoil}
\begin{figure*}[!htb]
 \centering
 \begin{subfigure}[b]{0.32\textwidth}
  \includegraphics[width=\textwidth]{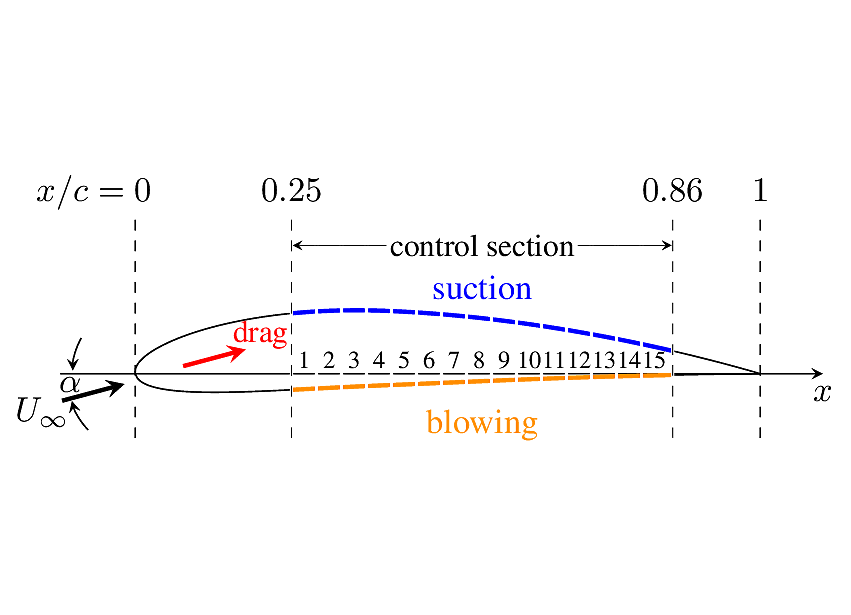}
  \caption{}
  \label{fig:profile}
 \end{subfigure}
 \hfill
 \begin{subfigure}[b]{0.32\textwidth}
  \includegraphics[width=\textwidth]{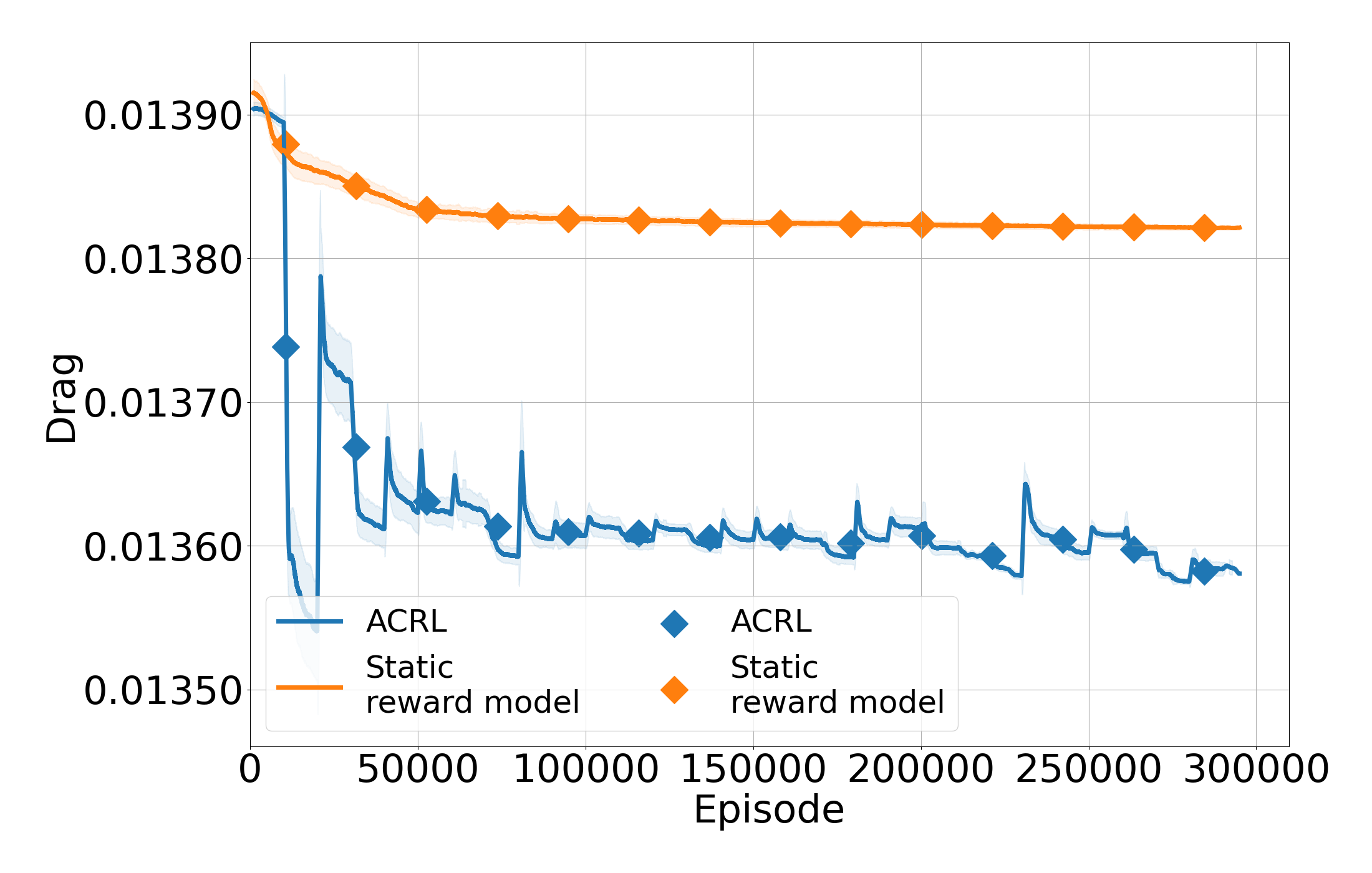}
  \caption{}
  \label{fig:cfd_training}
 \end{subfigure}
 \hfill
 \begin{subfigure}[b]{0.32\textwidth}
  \includegraphics[width=\textwidth]{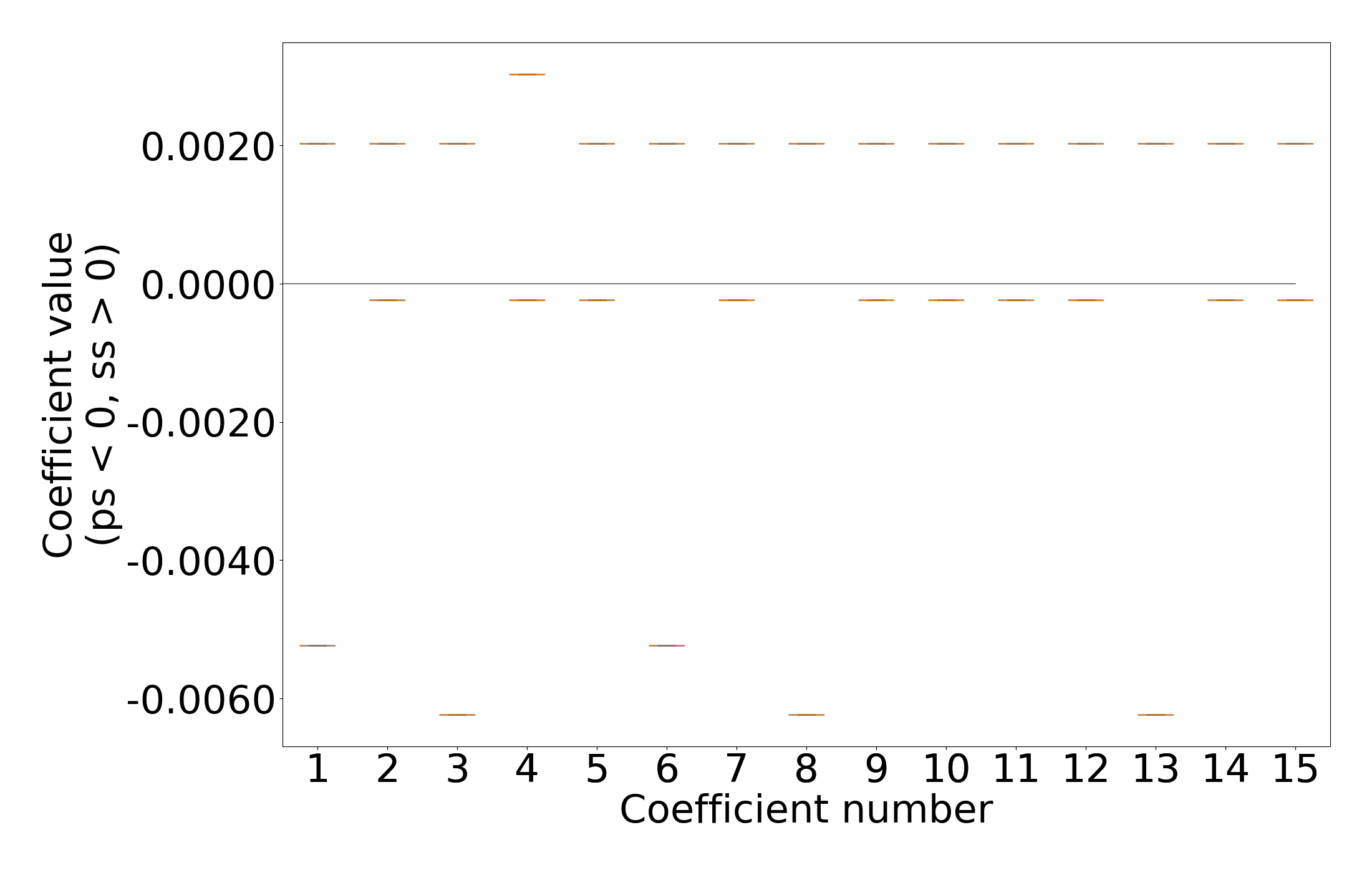}
  \caption{}
  \label{fig:cfd_profiles}
 \end{subfigure}
 \caption{
  (a) blowing/suction distribution discretized with 30 coefficients
  corresponding to 15 sections on each side of the considered airfoil.
  (b) drag evolution of two independent runs.
  (c) coefficient distribution for low-drag profiles.
 }
\end{figure*}

Our ACRL method is applicable to a large number of different tasks in
natural sciences and engineering, not only limited to chemistry.
Therefore, in this section we present the results of a task in engineering,
namely the reduction of airflow drag around an airfoil, e.g.\ an airplane
wing (see Section~\ref{sec:applications}).
The objective in this task was to find a set of coefficients minimizing drag
and to analyze the resulting profiles.
Figure~\ref{fig:cfd_training} shows the evolution of drag during $300000$
episodes of training.
The discrete jumps of the ACRL model coincide with retraining of the reward
model every $10000$ episodes.
As higher mean constraints are highly correlated with lower drag, we
choose samples for ground-truth evaluation based on reward rather
than drag.
Dots represent oracle-based ground-truth evaluations of random profiles
sampled during training.

The results demonstrate that the ACRL agent is able to find profiles with
significantly lower drag coefficients than the static reward model.
They also show that in this task (in contrast to the molecular improvement
task) it is crucial to actively update the reward model during training.
This is related to the fact that in order to improve upon the initially
uniform profile, the RL agent has to perform a constrained optimization in
high-dimensional real space (30-dimensional in our case).
Accurate reward model predictions require sufficient coverage of the
relevant space within the initial dataset which is difficult to assert
because the relevant region is, in general, not known, which is also true
for many other real-world problems.
As a consequence, an agent trained without active updates of the reward
model only slightly improves upon a uniform profile.
At the same time, model updates result in sharp drops of both predicted and
ground-truth drag especially in the beginning of training as the relative
effect of new ground-truth samples is high and the RL agent probably
exploits wrong predictions of the early-stage reward models.
This effect decreases as more and more samples are obtained along the
trajectories towards the low-drag region in parameter space.

\begin{table}[!htb]
 \centering
 \begin{tabular}{||c c c c||}
  \hline
  Task        & Oracle queries & Model queries   & Speed-up    \\
  \hline\hline
  Mol.\ opt.\  & $\sim$4,000    & $\sim$200,000   & $\sim$50    \\
  \hline
  Mol.\ imp.\  & $\sim$4,000    & $\sim$25,000    & $\sim$6.25  \\
  \hline
  Drag opt.\   & $\sim$3,000    & $\sim$9,000,000 & $\sim$3,000 \\
  \hline
 \end{tabular}
 \caption{
  Relative speed-up factors for different tasks comparing the number of
  oracle and model queries.
 }
 \label{tab:speedup}
\end{table}

Figure~\ref{fig:cfd_profiles} shows the distribution of a small number of
low-drag profiles sampled with ground-truth labels during training.
The resulting profiles are non-trivial and have a regular, alternating
pattern of coefficients with physically explainable
meaning~\cite[]{kametani2016drag,mahfoze2019reducing,stroh2016global}.
We note that due to various limitations of the simulated environment such as
discretization of action space, a limited number of coefficients due to
limitations in OpenFOAM simulations (used as an oracle) and limited episode
length, these results are only locally optimal w.r.t.\ our setup.
Yet, we find consistent, physically interpretable and highly non-trivial
results.

 \section{Conclusion}\label{sec:conclusion}
We introduced ACRL, an extension to standard reinforcement learning methods
in the context of (computationally) expensive rewards, which models the
reward of given applications using machine learning models.
Because optimal regions in the search spaces are not known a priori and thus
typically not included in initial training sets, we use active learning
during the exploration of the state space to update the reward model.
This way, the number of reward evaluations can be significantly reduced,
leading to a speed-up of up to 3000 in the presented applications.
This reduction enables the application of reinforcement learning algorithms,
in our case Double DQN, to real-world problems where reward evaluation can
become prohibitively expensive, and where conventional optimization methods
cannot be applied due to dynamic constraints which require generalization
across multiple problem instances.
We show in three applications - one for benchmarking purposes and two real
applications with expensive rewards - that training an agent with modelled
rewards yields reliable results and non-trivial solutions in complex
environments.
The transferability and wide applicability of our approach paves way for the
exploration of a wide range of real-world application areas using
reinforcement learning methods.

 \textbf{Acknowledgements}\\
 We would like to thank the Federal Ministry of Economics and Energy under
 Grant No. KK5139001AP0.
 We acknowledge support by the Federal Ministry of Education and Research
 (BMBF) Grant No. 01DM21001B (German-Canadian Materials Acceleration Center).
 We acknowledge funding by the German Research Foundation (Deutsche
 Forschungsgemeinschaft, DFG) within Priority Programme SPP 2331.

 \bibliographystyle{unsrt}
 \bibliography{ms}

 \section{Supplementary Information}\label{sec:supplementary-information}

\subsection*{Comparison of querying strategies for retraining}

\begin{figure*}
 \begin{subfigure}{.5\textwidth}
  \centering
  \includegraphics[width=0.8\textwidth]{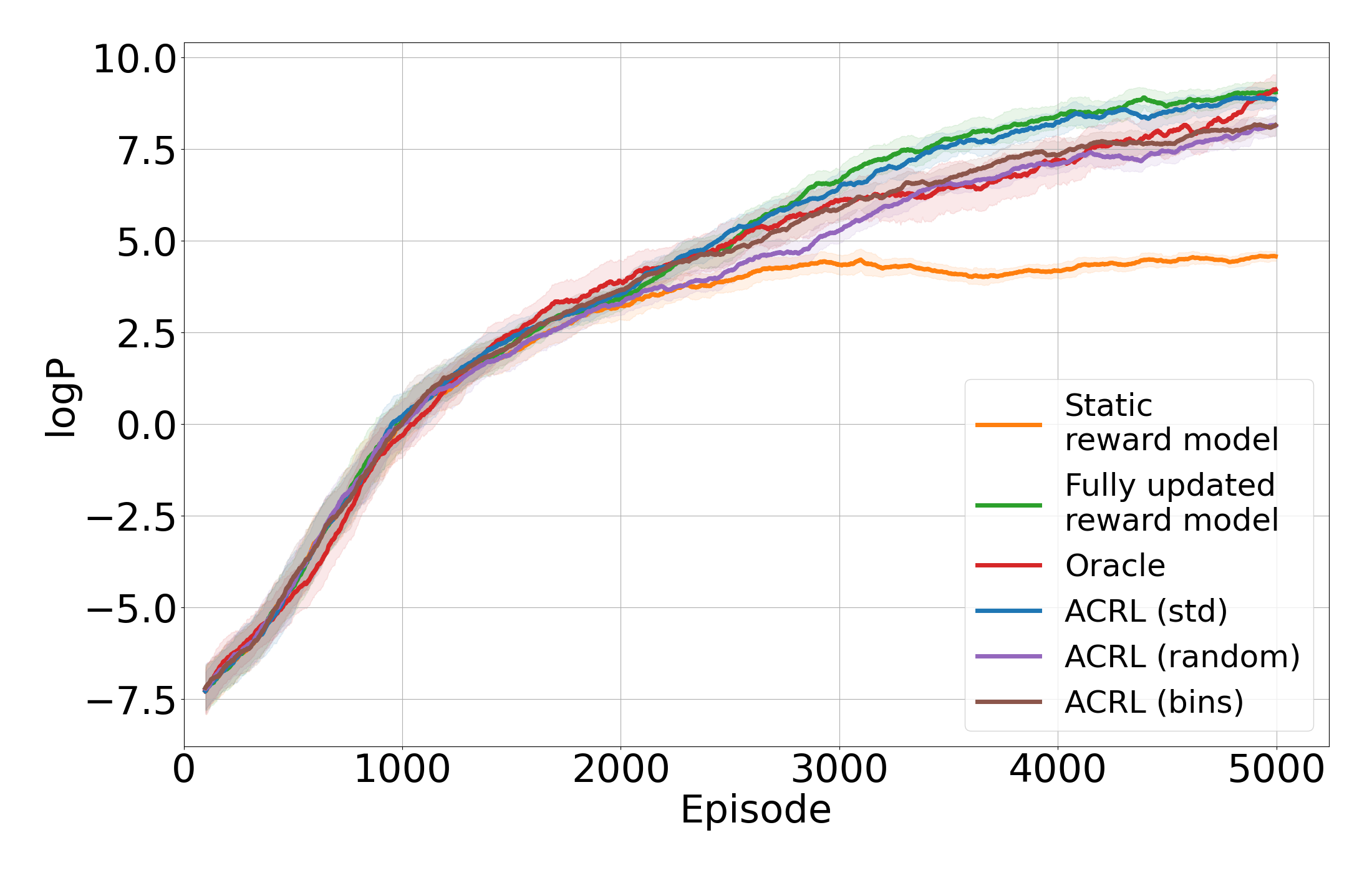}
  \caption{logp}
  \label{fig6:logp}
 \end{subfigure}%
 \begin{subfigure}{.5\textwidth}
  \centering
  \includegraphics[width=0.8\textwidth]{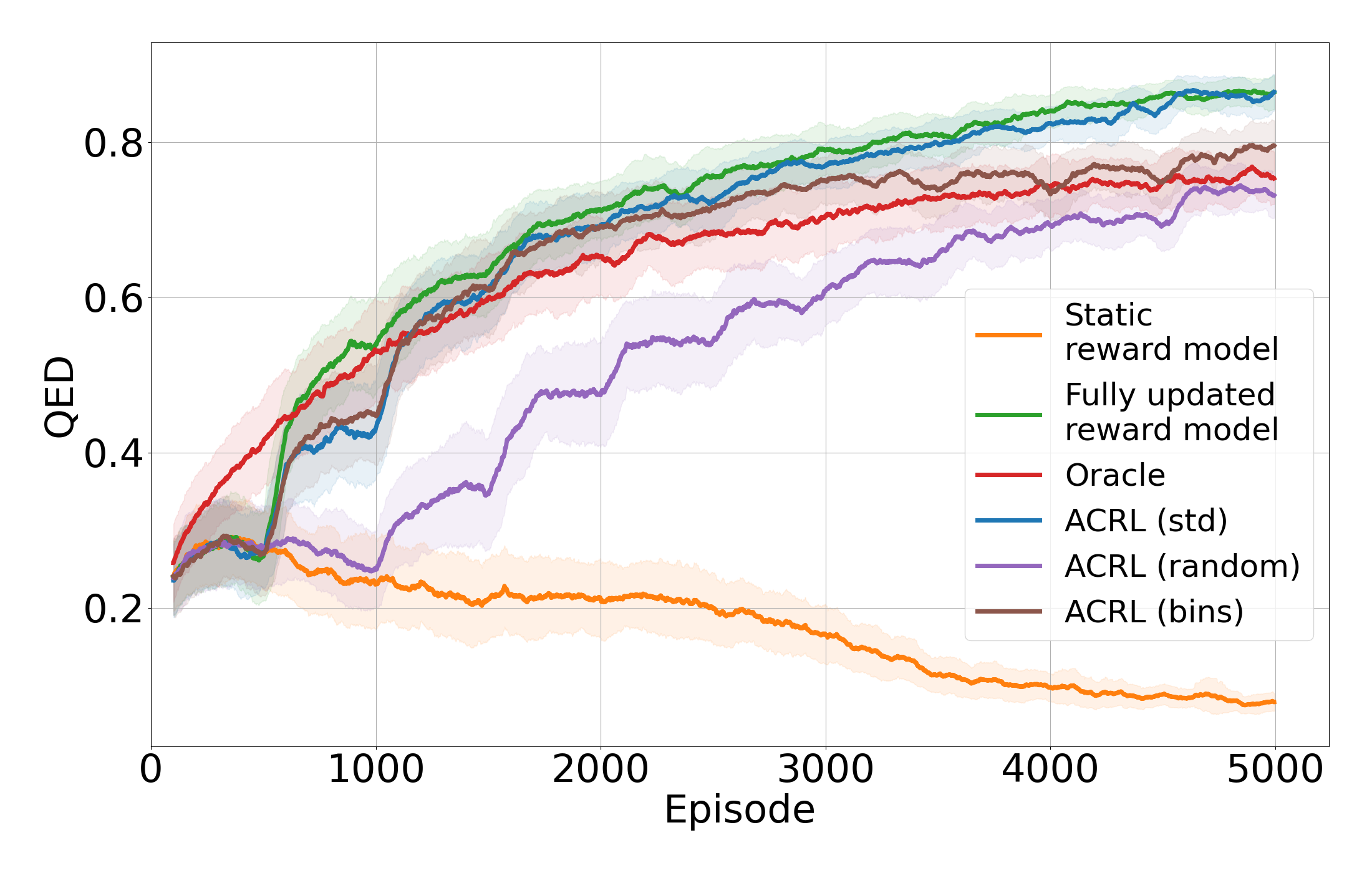}
  \caption{QED}
  \label{fig6:qed}
 \end{subfigure}
 \caption{Comparison of different selection modes, random (red), standard
 deviation based (yellow) and bin-based (purple) on 800 sampled points.}
 \label{fig:fig6}
\end{figure*}

The three strategies for selection of points presented were compared on a
constant number of selected points (800) (Figure~\ref{fig:fig6}).
In each case, three reward models are initially trained with three different
train-test splits of the original QM9 dataset and used for prediction later on.
The mean of the three reward models' predictions is used as a final reward
for the agent to maximize, and the standard deviation of these are
calculated to give an idea of the uncertainty on prediction.
Based on these models, three selection modes are studied.
In the first setting, the models are retrained by randomly sampling a number
of points from the initial QM9 test set as well as newly generated points
during the learning process.
In a second setting, the points with the highest standard deviations of
model predictions are sampled and the models are updated using these points
as train set.
This is based on the assumption that the points with the highest standard
deviations of model predictions (points in which the three models
"disagree") are more likely to come from outside the original train
distribution, thus potentially representing the points that the model needs
to learn from in order to improve its predictions during the agent training
process.
In this sense, having three reward models instead of only one could provide
a good basis for the selection of points.
Finally, a third approach in sampling points is based on classifying the
previously obtained test set and newly generated points in different bins
before selecting points with the highest standard deviation (of model
predictions).
This stems from the fact that in certain situations, points with the highest
standard deviations could represent outliers, and therefore could prevent
the reward models from learning the main trend of the data.
The strategy of bin-based selection offers a solution to this by selecting
points with the highest standard deviations while respecting the initial
data distribution.
It is important to note that all these sampling processes are done
separately for each model to be retrained, since their respective starting
train and test sets are not the same to start with.
Therefore, the three models will never be updated on the same training data,
and will thus provide independent predictions, depending on the points they
were trained on.
Overall, the standard deviation based sampling method performed best and was
thus used in the main part of this paper.

\subsection*{Comparison of the number of points selected for retraining}
\begin{figure*}
 \begin{subfigure}{.5\textwidth}
  \centering
  \includegraphics[width=0.8\textwidth]{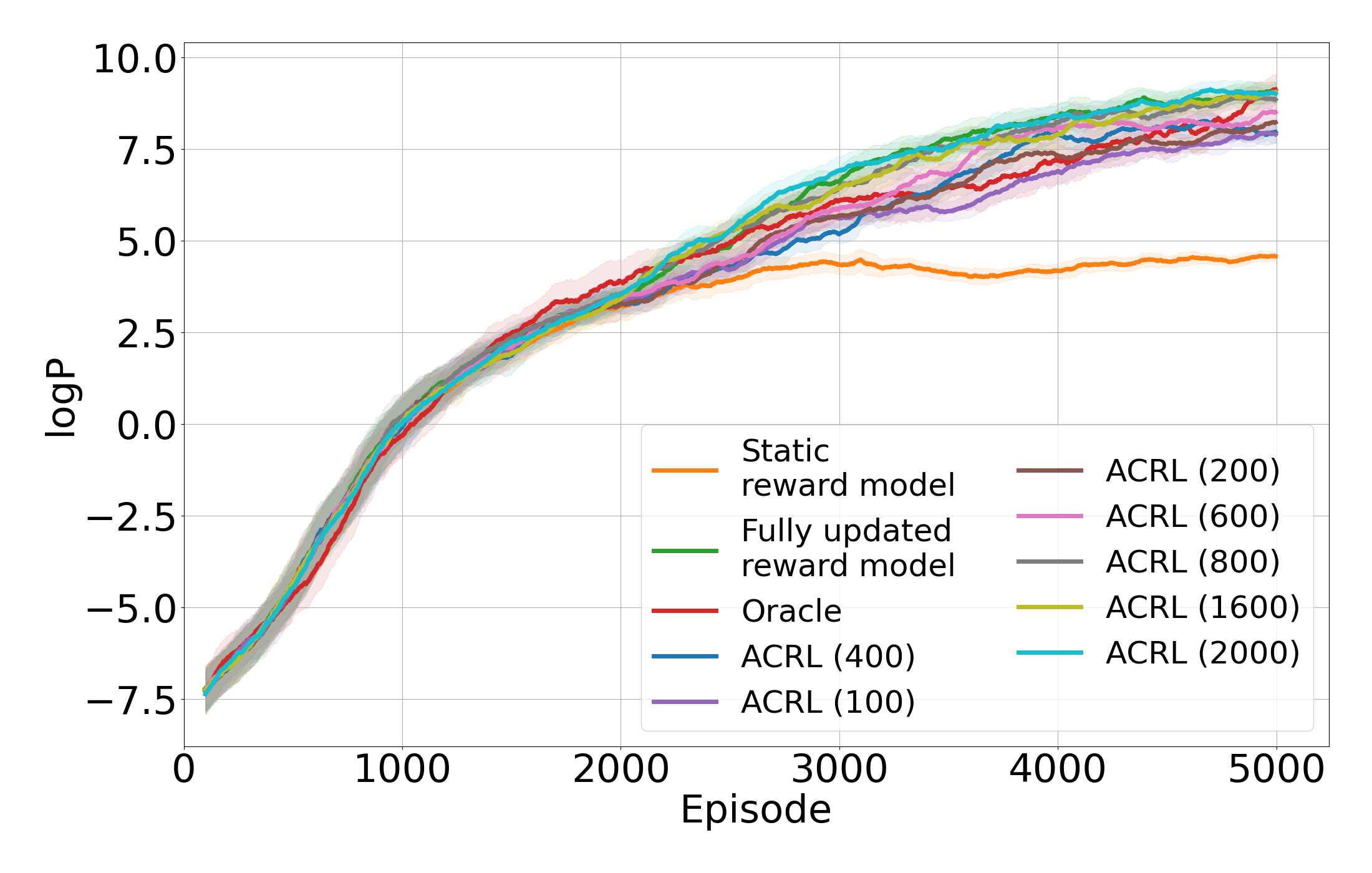}
  \caption{logp}
  \label{fig7:logp}
 \end{subfigure}%
 \begin{subfigure}{.5\textwidth}
  \centering
  \includegraphics[width=0.8\textwidth]{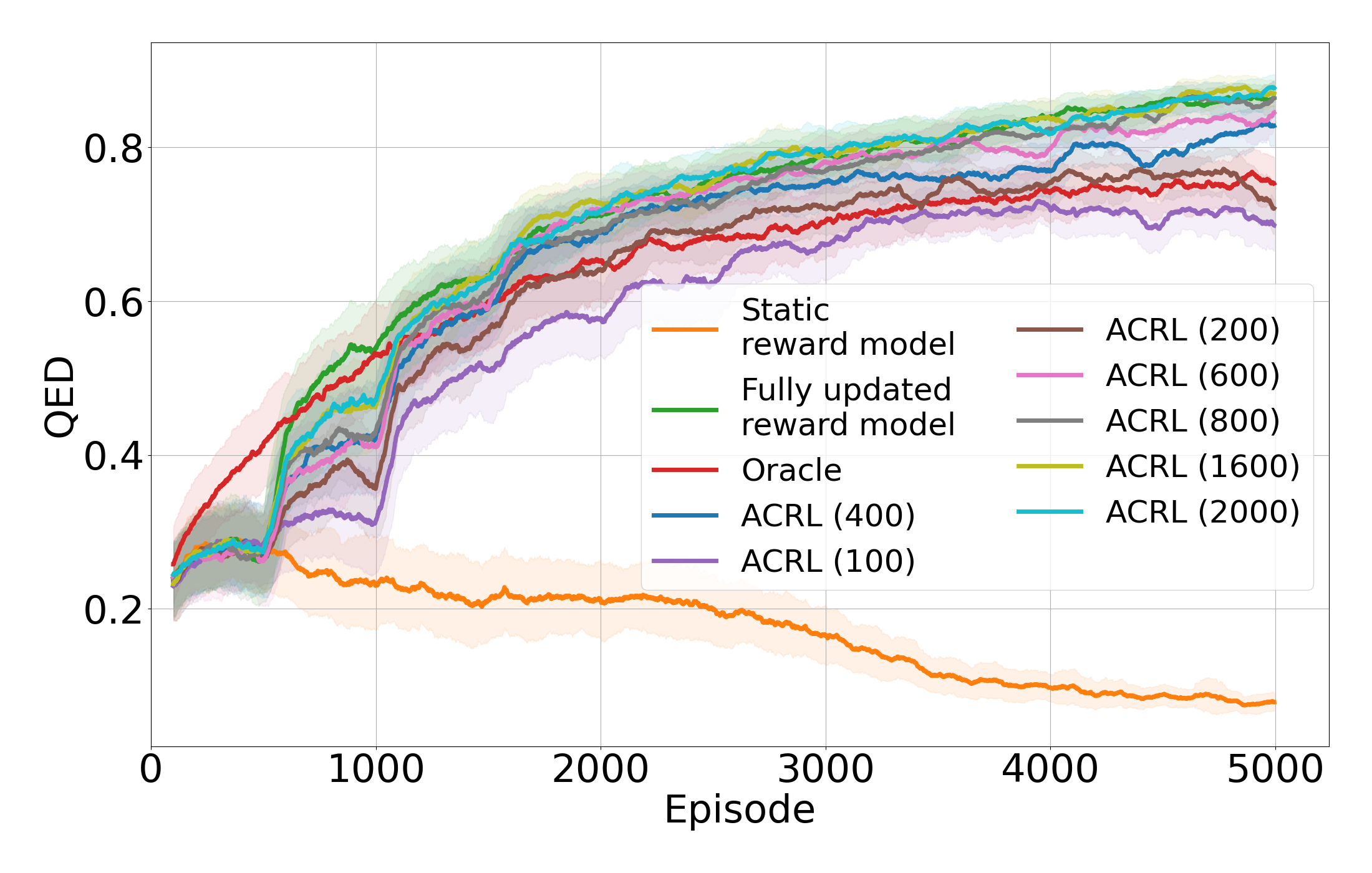}
  \caption{QED}
  \label{fig7:qed}
 \end{subfigure}
 \caption{Comparison of reward models retrained on a varying number of
 points selected based on standard deviation}
 \label{fig:fig7}
\end{figure*}

Given the best strategy (standard deviation based), the effect of the number
of points selected for reward model retraining was studied
(Figure~\ref{fig:fig7}).
The minimal number of points that was consistently comparable to the real
reward was 400 points.

\subsection*{Study of the effect of varying degrees of randomness on learning}
\begin{figure*}
 \begin{subfigure}{.5\textwidth}
  \centering
  \includegraphics[width=0.8\textwidth]{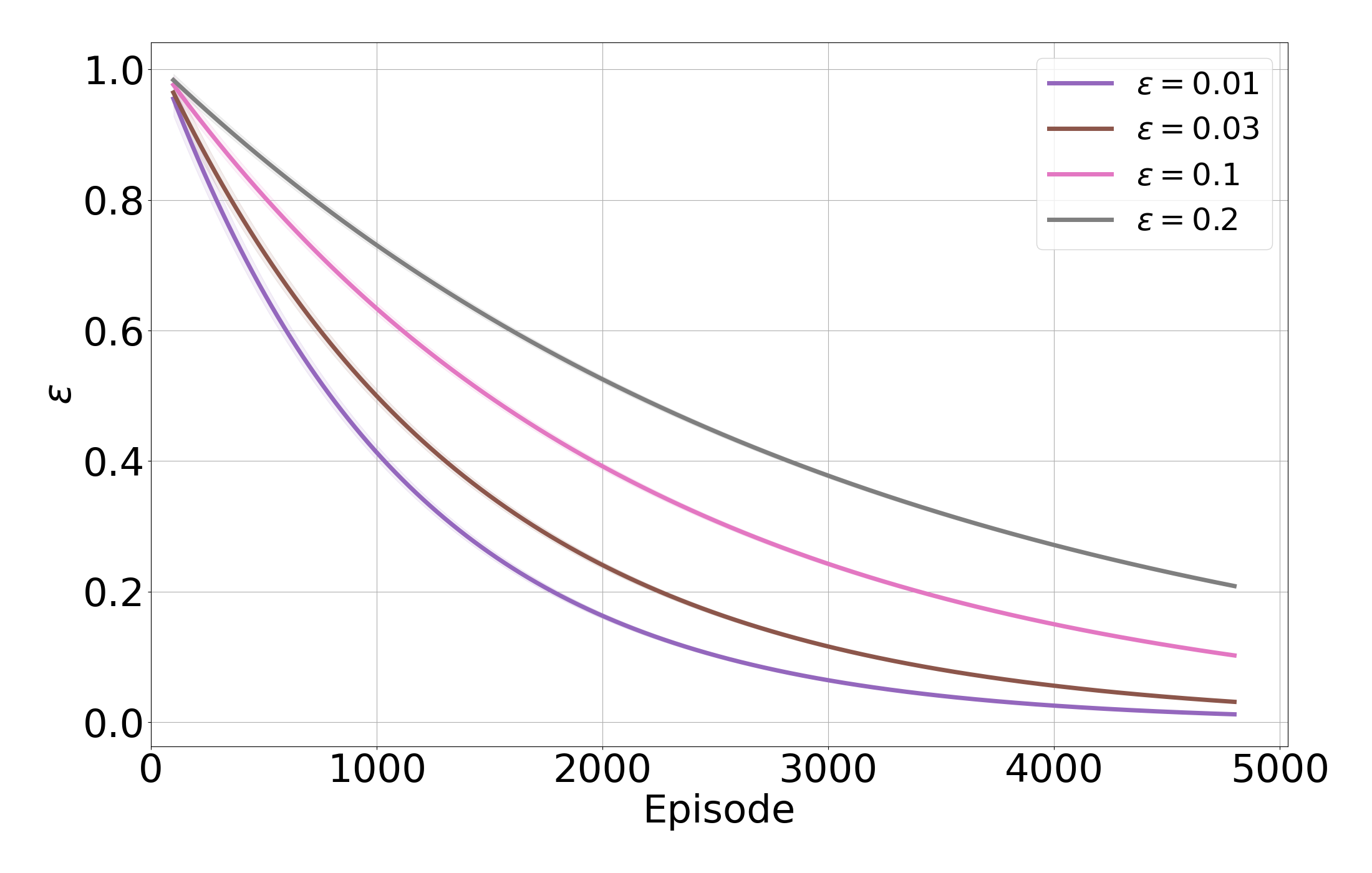}
  \caption{$\epsilon$-values' exponential decay with varying end points}
  \label{fig8:randomness}
 \end{subfigure}%
 \begin{subfigure}{.5\textwidth}
  \centering
  \includegraphics[width=0.8\textwidth]{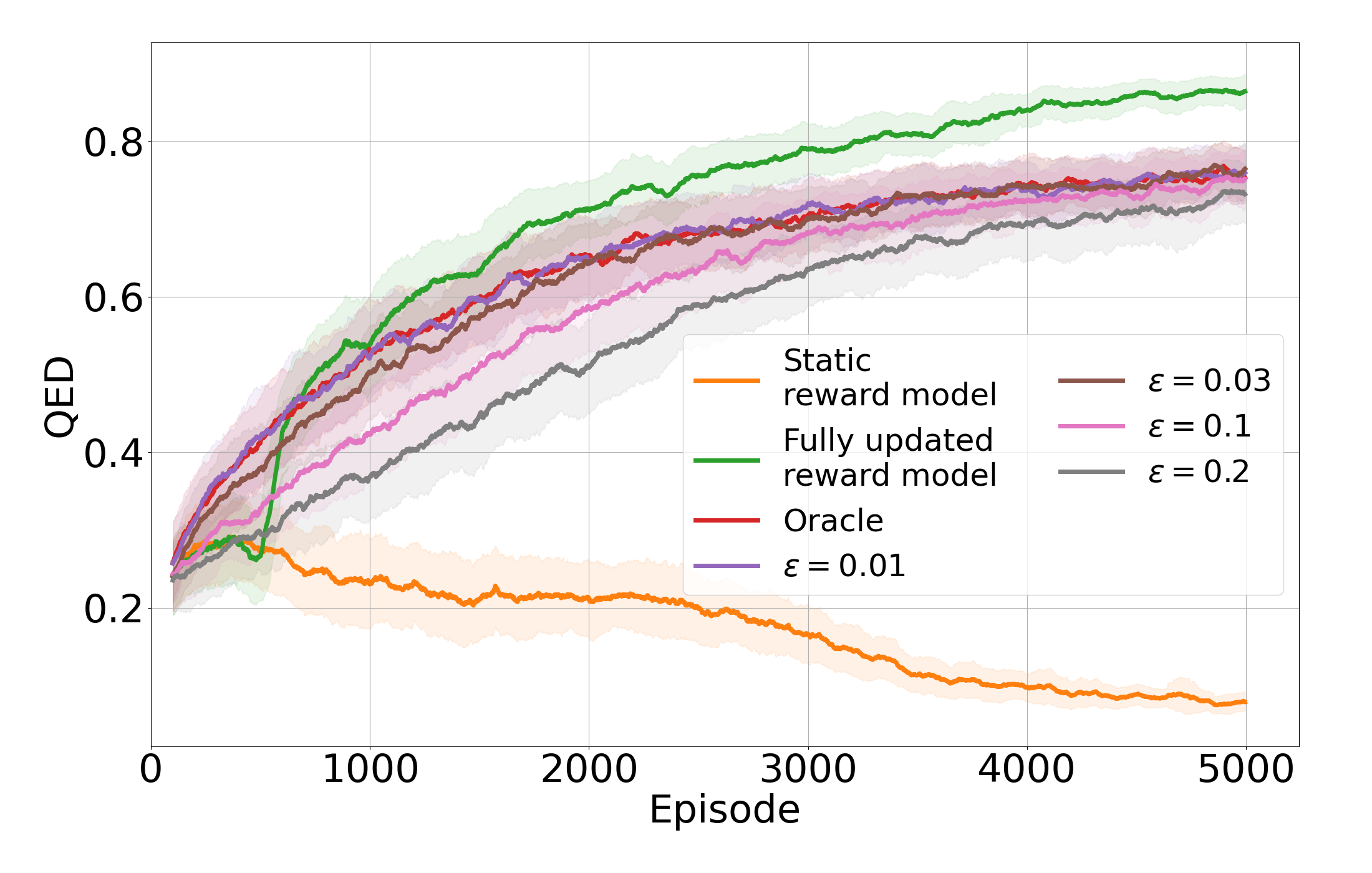}
  \caption{Results on the QED task}
  \label{fig8:results}
 \end{subfigure}
 \caption{The effect of increasing randomness by reaching different
  $\epsilon$ end values with the same exponential decay function.}
 \label{fig:fig8}
\end{figure*}

The $\epsilon$ values in the MolDQN paper start with values of 100\% and
decrease exponentially at each episode until they reach a percentage of 1\%
at episode 5000.
In this study, our aim was to compare the effect of the $\epsilon$ end value
(last value of randomness) on the real reward agent at episode 4800,
considering that the agent does not choose any more random actions from
episode 4800 to 5000 (to better guide it towards the end goal).
Results are shown in Figure~\ref{fig:fig8}.
We concluded that increasing the $\epsilon$ end values (increasing
randomness, with the hope of favoring exploration) did not help the agent
reach better rewards.

\subsection*{Study of the effect of the decay function form}
\begin{figure*}
 \begin{subfigure}{.5\textwidth}
  \centering
  \includegraphics[width=0.8\textwidth]{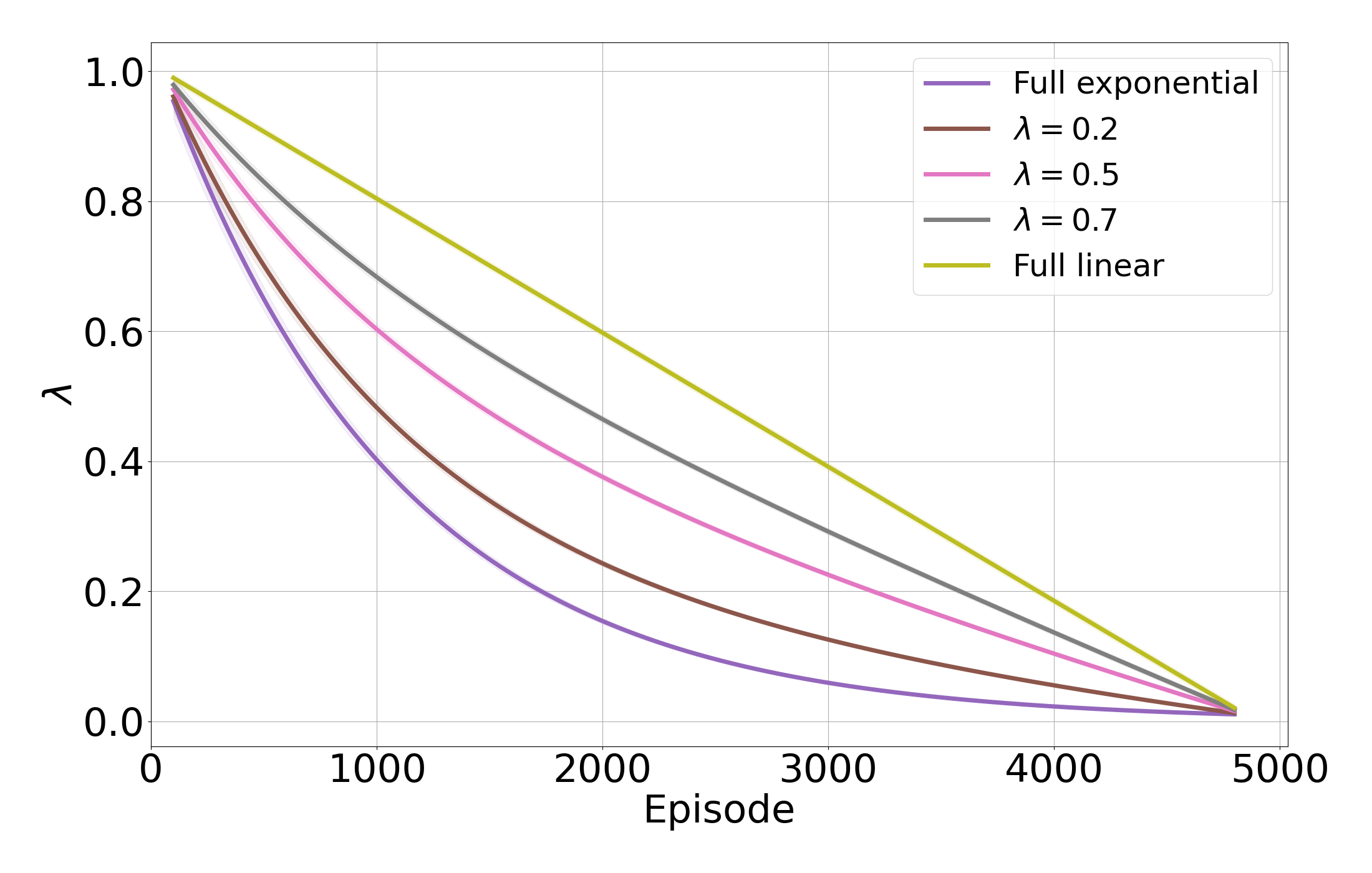}
  \caption{Different forms of $\epsilon$-decay functions}
  \label{fig9:function}
 \end{subfigure}%
 \begin{subfigure}{.5\textwidth}
  \centering
  \includegraphics[width=0.8\textwidth]{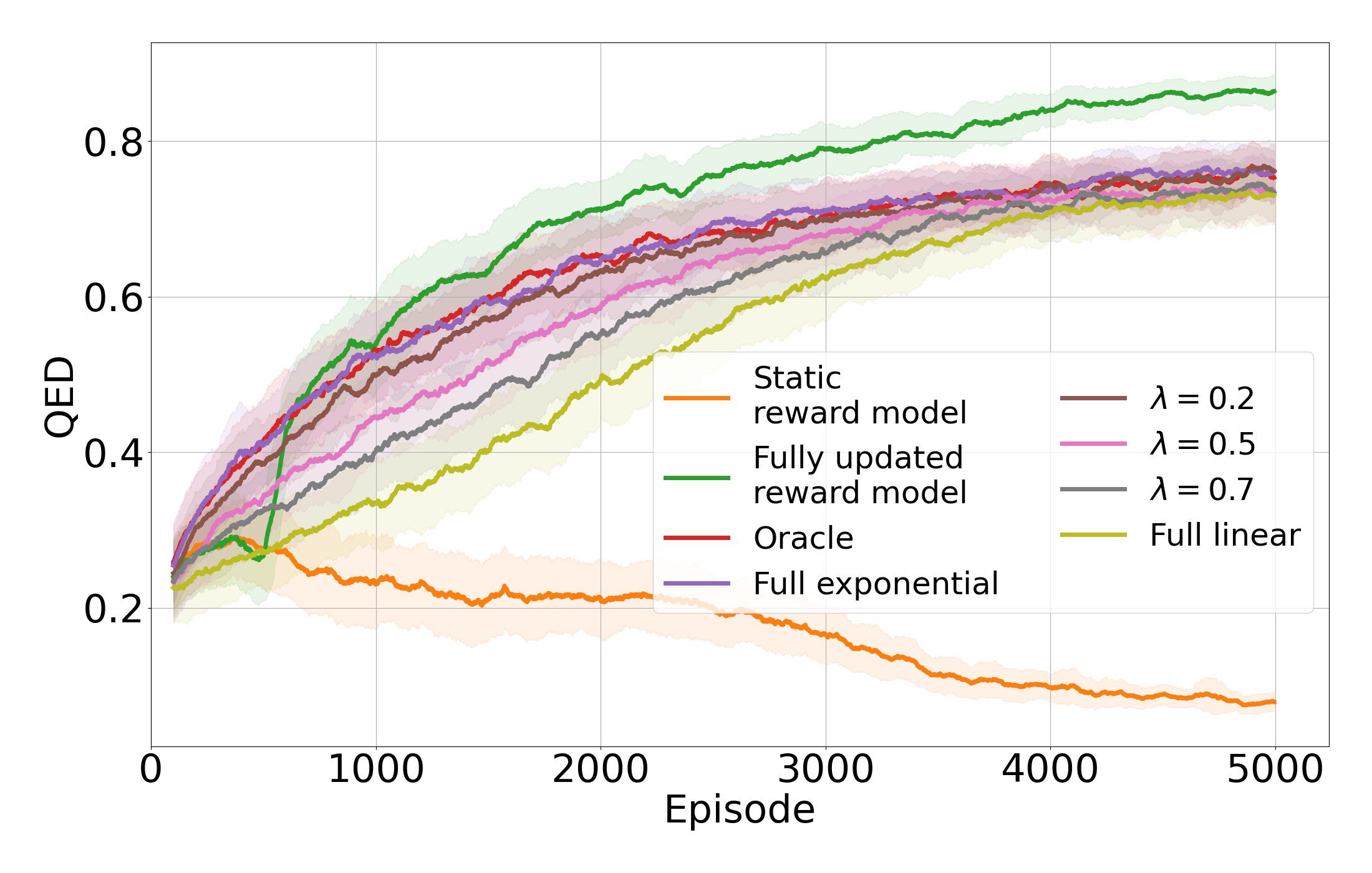}
  \caption{Results on the QED task}
  \label{fig9:results}
 \end{subfigure}
 \caption{The effect of different $\epsilon$-decay functions with the same
 randomness end value.}
 \label{fig:fig9}
\end{figure*}

The $\epsilon$-decay function used in MolDQN is an exponential decay function.
In this study, we choose to study the effect of varying decay function forms
by adding a linear component to the exponential function with varying
fractions.
The equation is the following:
\begin{equation}
 \epsilon(t) = \epsilon_0(\lambda(1-\beta t) + (1-\lambda)\alpha^t)
 \label{eq:equation2}
\end{equation}
with $1-\beta t$ and $\alpha^t$ the linear and exponential components
respectively, $\epsilon_0$ the starting randomness (at 100\%),
$\beta$ and $\alpha$ constants that depend on the starting and end values
(we choose an end value of 1\% at episode 4800), and $\lambda$
the fraction (relative importance) of the linear component.
At a $\lambda$ of 0, the decay-function is fully exponential, and at a
$\lambda$ of 1, it is fully linear.
Results are shown in Figure~\ref{fig:fig9}.
We conclude here that a fully exponential function is more convenient for
the learning of the agent.

\subsection*{Comparison of different mean constraints}
\begin{figure*}[!htb]
 \centering
 \begin{subfigure}[b]{0.49\textwidth}
  \includegraphics[width=\textwidth]{figures/main_cfd_small_range_training}
  \caption{}
  \label{fig:si_cfd_small_training}
 \end{subfigure}
 \begin{subfigure}[b]{0.49\textwidth}
  \includegraphics[width=\textwidth]{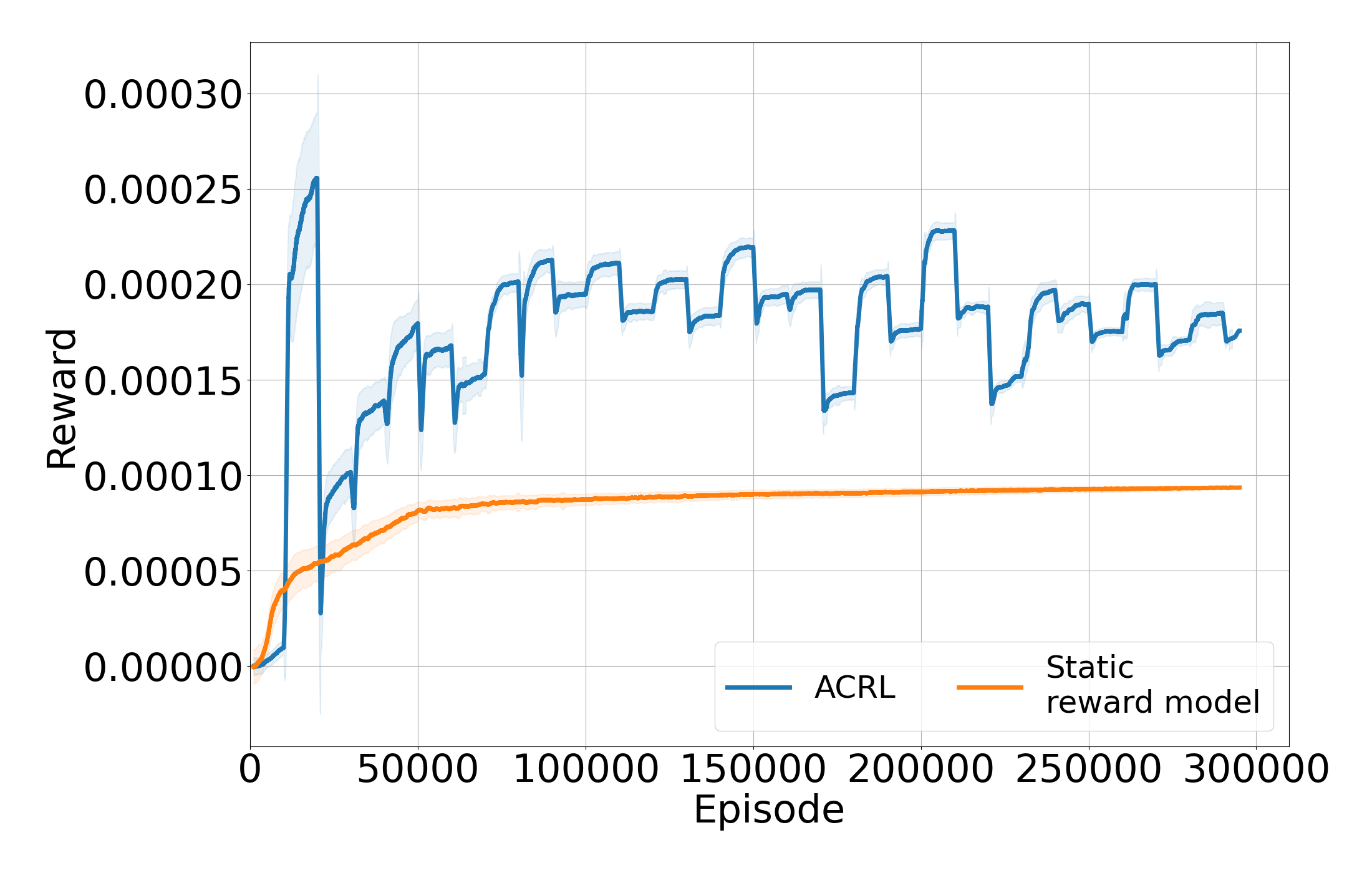}
  \caption{}
  \label{fig:si_cfd_small_reward}
 \end{subfigure}
 \hfill
 \caption{
  Results for different mean constraints in $[0.0019,0.0021]$.
 }
 \label{fig:cfd_constraints_small}
\end{figure*}

\begin{figure*}[!htb]
 \centering
 \begin{subfigure}[b]{0.49\textwidth}
  \includegraphics[width=\textwidth]{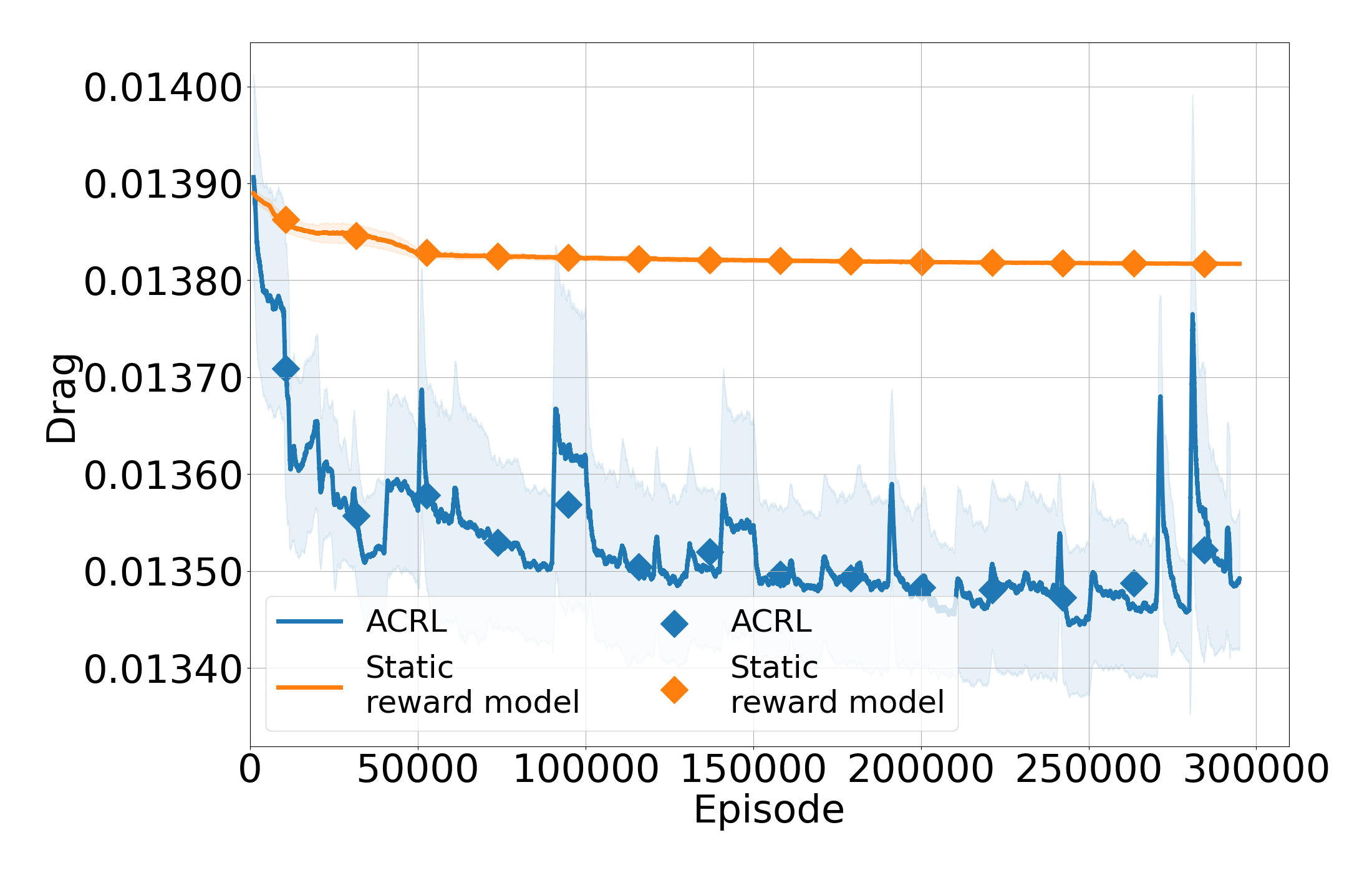}
  \caption{}
  \label{fig:si_cfd_large_training}
 \end{subfigure}
 \begin{subfigure}[b]{0.49\textwidth}
  \includegraphics[width=\textwidth]{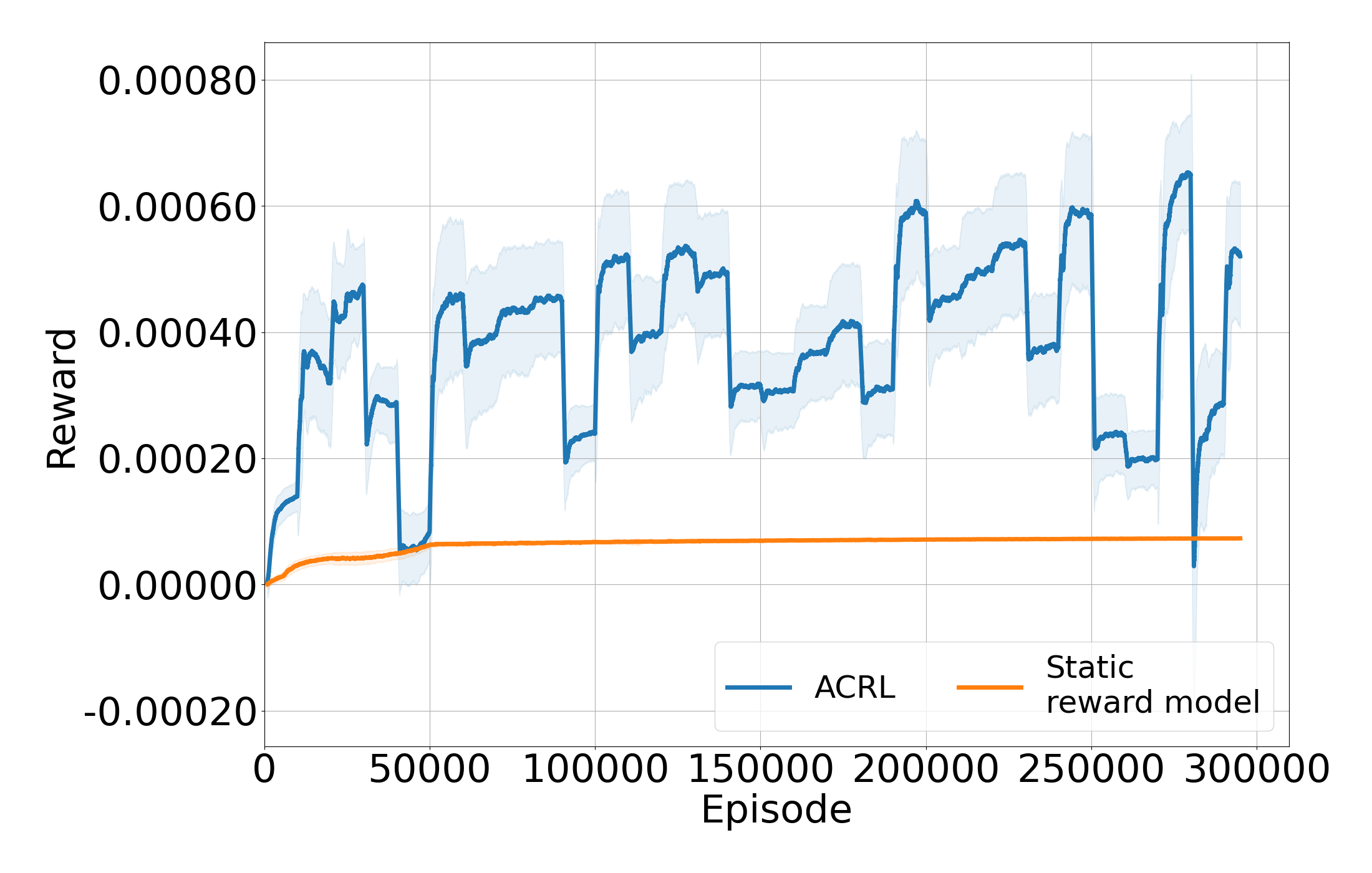}
  \caption{}
  \label{fig:si_cfd_large_reward}
 \end{subfigure}
 \hfill
 \caption{
  Results for different mean constraints in $[0.0015,0.0025]$.
 }
 \label{fig:cfd_constraints_large}
\end{figure*}

In this section, we present a more detailed description of the results for
the drag optimization task.
Our initial training dataset contains 5000 random profiles with a mean value
centered around $\pm0.002$ on each side.
For a constraint configuration matching the distribution of the initial
dataset, Figures~\ref{fig:si_cfd_small_training}
and~\ref{fig:si_cfd_small_reward} show the evolution of drag and reward,
respectively.
In order to test extrapolation and generalization capabilities of our ACRL
method, we repeated the same experiment with a larger constraint interval
which lies outside the initial training distribution.
The results in Figures~\ref{fig:si_cfd_large_training}
and~\ref{fig:si_cfd_large_reward} show that ACRL is able to explore the
underrepresented space well, in contrast to a static reward model which
fails to guide the agent to explore the relevant solution spaces.
The higher variance stems from the fact that higher mean values (or
equivalently, higher total volume) trivially reduce drag.
Thus, in this experiment the agent encounters a higher diversity of states
in terms of their constraint.
Even though most of the observed states lie outside the initial training
distribution, an ACRL agent is still able to explore the relevant low-drag
space.
The importance of actively updating the reward model during training is
reflected by the results of agents using a static reward model in both
experiments.
Both agents trained with a static reward model achieve very similar results
in terms of drag, even though drag distributions vary considerably between
the experiments.
Only the ACRL agents are able to capture the variance of drag well, which is
especially high in Figure~\ref{fig:cfd_constraints_large} due a larger
constraint interval.
In contrast, static models fail to move outside their initially modelled
distribution, even though they predict drag values of encountered states
very accurately.

\end{document}